\documentclass[10pt,twocolumn,letterpaper]{article}
\usepackage[T1]{fontenc}

\usepackage[pagenumbers]{cvpr}

\usepackage{amsthm,mathtools}
\usepackage{multirow,array}
\usepackage{pgfplots}
\pgfplotsset{compat=1.18}
\usetikzlibrary{arrows.meta,patterns}

\definecolor{linkblue}{rgb}{0.21,0.49,0.74}
\usepackage[pagebackref,breaklinks,colorlinks,allcolors=linkblue,hyperfootnotes=false]{hyperref}
\hypersetup{
  pdftitle={Sharp First-Hit Identification from Finite Depth Samples},
  pdfauthor={Hamidreza Hasani Balyani, Mahtab Movahhedrad, Seyed Pouyan Mousavi Davoudi}
}

\newcounter{algorithm}
\newtheorem{theorem}{Theorem}
\newtheorem{proposition}[theorem]{Proposition}
\newtheorem{lemma}[theorem]{Lemma}
\newtheorem{corollary}[theorem]{Corollary}

\newtheorem{remark}[theorem]{Remark}
\newcommand{\R}{\mathbb R}
\newcommand{\Fcal}{\mathcal F}
\newcommand{\Qcal}{\mathcal Q}
\newcommand{\Zcal}{\mathcal Z}
\newcommand{\lo}{\underline f}
\newcommand{\up}{\overline f}

\newcommand{\diam}{\operatorname{diam}}
\newcommand{\Lip}{\operatorname{Lip}}
\newcommand{\norm}[1]{\left\lVert#1\right\rVert}

\newcommand{\eps}{\varepsilon}

\newcommand{\EtwoRecords}{4,900}
\newcommand{\EtwoMaxRel}{$1.2\times10^{-6}$}
\newcommand{\EtwoMaxRelDAv}{$1.2\times10^{-6}$}
\newcommand{\EtwoMaxRelZoe}{$1.0\times10^{-6}$}
\newcommand{\OoneQueries}{810,000}
\newcommand{\OoneMembers}{564,894}
\newcommand{\OtwoRecords}{51,660}
\newcommand{\OtwoConditions}{21}
\newcommand{\ArchiveRecords}{3,175,200}
\newcommand{\ArchiveMembers}{553,215/553,215}
\newcommand{\ArchiveChecks}{80}
\newcommand{\OoneJointChecks}{30,981}
\newcommand{\OoneJointFailures}{0}
\newcommand{\FreiburgText}{The study evaluated 4 scene-disjoint test sequences (primary band, the calibration sequences' pointwise 95th percentile of relative inverse-depth error, and primary slope variant): Depth Anything V2 (band $\pm$59\%): 315 of 6,912 queries feasible, measured containment 87.32\% (equal-sequence mean over 3 sequences); ZoeDepth (band $\pm$85\%): 318 of 6,912 queries feasible, measured containment 100.00\% (equal-sequence mean over 3 sequences); Depth Pro (band $\pm$82\%): 318 of 6,912 queries feasible, measured containment 95.77\% (equal-sequence mean over 3 sequences). In 1 of the 4 sequences no model produced a feasible output. Containment is not uniform: on the \texttt{teddy} sequence Depth Anything V2 contains 44 of 71 and Depth Pro contains 62 of 71 measured targets. No query was accepted at 5\% by any model, consistent with bands this wide; the Freiburg study therefore says nothing about accepted-output error.}
\newcommand{\DepthProText}{On the 10 held-out scenes and the 4,900 primary queries, Depth Pro sources are in class for 10.12\% of queries (equal-scene mean) and accepted at 5\% for 5.89\% [2.45, 9.88]. All 565 digital class members are contained (565/565). Without conditioning on membership, as in \cref{tab:heldout-coverage}, digital coverage is 898/1,415 and measured coverage is 1,836/2,945. Of the 339 queries accepted at 5\% with a measured reference, 330 have measured error above 5\% (equal-scene rate 76.88\% [50.00, 100.00] over the 8 scenes with such a query, against 100.0\% and 89.9\% for Depth Anything V2 and ZoeDepth in \cref{tab:heldout-bootstrap}); all 154 accepted queries with a digital reference have digital error above 5\%. Acceptance by itself therefore does not bound error for Depth Pro either. The run recorded 0 numerical errors.}
\newcommand{\OtwoWText}{O2-w ran all 21 conditions; 9 of 10 scenes were evaluable. Under the correct pose, 100.00\% of 2,700 queries were applicable and measured containment was 100.00\%. Pooled measured containment falls to 82.55\% under 0.5\textdegree{} rotations and 87.26\% under 0.05$B$ translations (lowest single condition: pitch $+0.5$\textdegree{}, 71.22\% equal-scene mean). Acceptance drops most under yaw $-0.5$\textdegree{} (38.33\%). Across all conditions, 0 of 51,345 accepted queries had measured error above 5\%.}
\newcommand{\OptionalResultsRows}{%
O1 primary cell & 3,000 & 98.40\% & 2,382/2,382 & 98.40 [96.20, 100.00] & 0/2,952 \\
O1 sensitivity settings & 807,000 & 96.20\% & 562,512/562,512 & 80.92 [75.73, 85.78] & 0/653,035 \\
O2 correct pose & 2,460 & 0.00\% & -- & 0.00 & -- \\
O2 20 perturbations & 49,200 & 0.00\% & -- & 0.00 & -- \\
O2-w correct pose & 2,700 & 100.00\% & -- & 100.00 & 0/2,700 \\
O2-w rotation 0.1\textdegree{} (4) & 10,800 & 100.00\% & -- & 100.00 & 0/10,800 \\
O2-w rotation 0.5\textdegree{} (4) & 10,800 & 59.86\% & -- & 59.86 & 0/6,465 \\
O2-w translation 0.01$B$ (6) & 16,200 & 100.00\% & -- & 100.00 & 0/16,200 \\
O2-w translation 0.05$B$ (6) & 16,200 & 93.70\% & -- & 93.70 & 0/15,180 \\
}

\title{Sharp First-Hit Identification from Finite Depth Samples}

\author{Hamidreza Hasani Balyani\\
Independent Researcher
\and
Mahtab Movahhedrad\\
University of Southern California
\and
Seyed Pouyan Mousavi Davoudi\\
Islamic Azad University, Amol Branch
}

\begin{document}
\maketitle
\begin{abstract}
Which depths in a new camera view are determined by finitely many source-depth
measurements? Pointwise uncertainty bounds do not answer this question:
unobserved geometry between samples can change which surface is encountered
first. We study a bounded Lipschitz class of continuous source inverse-depth
fields with interval observations and known camera pose. Building on classical
Lipschitz envelopes, we characterize the exact set of attainable first-hit
depths and construct a globally admissible scene for every value in that set.
The set can be disconnected. We also give a necessary and sufficient test
for several proposed depths to be first hits of one shared scene. An epipolar
reduction uses scalar cone inequalities without enumerating surfaces. In a rectified model with
source spacing $h$, interval half-width $\eps$, baseline $b$, and slope bound
$L$, we prove a sharp worst-case inverse-depth diameter of
$\max\{Lh+2\eps,2\eps/(1-bL)\}$ when $bL<1$, subject to sufficient depth and
domain range for attainment. Exact samples still give the sharp bound $Lh$
at $bL=1$, while above this threshold ambiguity need not vanish as sampling
becomes arbitrarily dense. Any positive measurement uncertainty can destroy
uniform stability at the critical threshold. We provide constructive examples,
an exact rational validation oracle, and a frozen ten-scene evaluation
using reference, stereo, and learned-depth observations. The experiments
separate solver correctness under the scene assumptions from failure of
those assumptions: fixed learned-depth bands often yield narrow but
incorrect sets. The results identify what a specified scene class permits;
they do not infer a continuum regularity or physical-scene guarantee from
sampled predictions.
\end{abstract}

{\let\thefootnote\relax\footnotetext{This work has been submitted to the IEEE for possible publication. Copyright may be transferred without notice, after which this version may no longer be accessible.}}
\vfill\newpage
\section{Introduction}
\label{sec:intro}
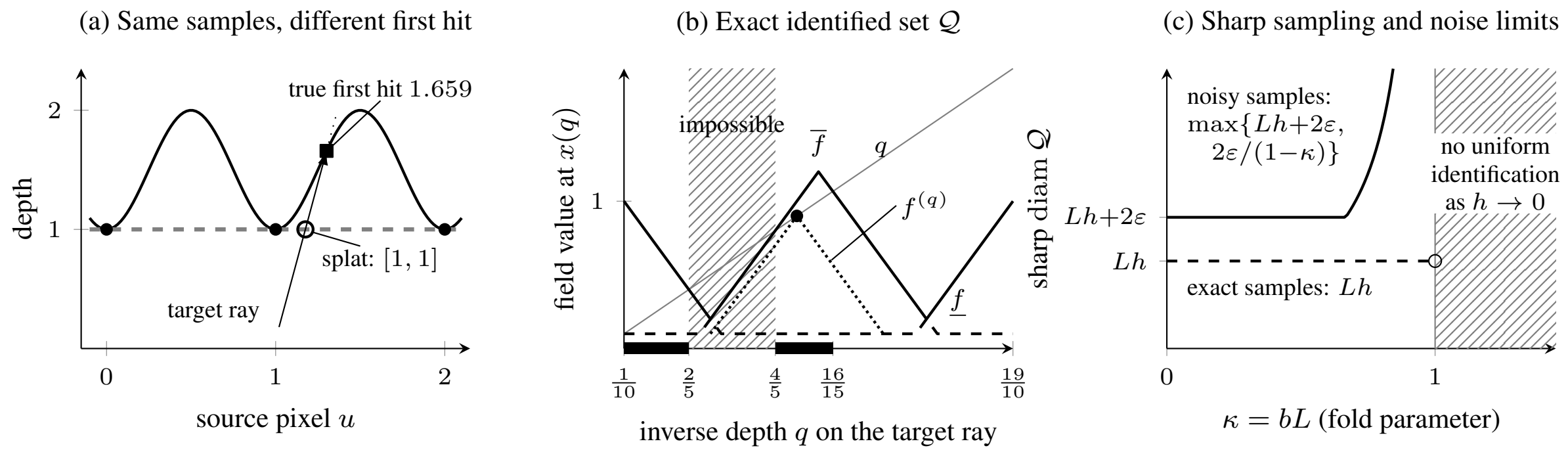
\begin{figure*}[t]
\centering
\pgfplotsset{
  teaser/.style={
    width=5.5cm, height=4.4cm, scale only axis=false,
    axis lines=left, tick label style={font=\scriptsize},
    label style={font=\footnotesize}, title style={font=\footnotesize, align=center},
    every axis plot/.append style={line join=round},
    clip=false,
  }
}
\begin{tikzpicture}
\begin{axis}[teaser, name=pa,
  xmin=-0.15, xmax=2.15, ymin=0.0, ymax=2.35,
  xtick={0,1,2}, ytick={1,2},
  xlabel={source pixel $u$}, ylabel={depth},
  ylabel style={yshift=-4pt},
  title={(a) Same samples, different first hit}]
  \addplot[black, thick, domain=-0.1:2.1, samples=120] {1+sin(deg(pi*x))^2};
  \addplot[gray, very thick, dashed, domain=-0.1:2.1] {1};
  \addplot[only marks, mark=*, mark size=1.6pt, black] coordinates {(0,1) (1,1) (2,1)};
  \addplot[black, thin, -{Latex[length=4pt]}] coordinates {(1.02,0.18) (1.301,1.659)};
  \addplot[black, thin, dotted] coordinates {(1.301,1.659) (1.36,1.97)};
  \addplot[only marks, mark=square*, mark size=1.8pt, black] coordinates {(1.301,1.659)};
  \addplot[only marks, mark=o, mark size=2.2pt, black, thick] coordinates {(1.176,1.0)};
  \node[font=\scriptsize, anchor=south, align=center, fill=white, inner sep=1pt] (hit) at (axis cs:1.62,2.08) {true first hit $1.659$};
  \draw[black, thin] (hit.south) -- (axis cs:1.32,1.68);
  \node[font=\scriptsize, anchor=north, fill=white, inner sep=1pt] at (axis cs:1.62,0.88) {splat: $[1,1]$};
  \draw[black, thin] (axis cs:1.5,0.86) -- (axis cs:1.21,0.98);
  \node[font=\scriptsize, anchor=east] at (axis cs:0.98,0.3) {target ray};
\end{axis}
\begin{axis}[teaser, name=pb, at={(pa.east)}, anchor=west, xshift=1.55cm,
  xmin=0.1, xmax=1.9, ymin=0.0, ymax=1.9,
  xtick={0.1,0.4,0.8,1.0667,1.9},
  xticklabels={$\tfrac1{10}$,$\tfrac25$,$\tfrac45$,$\tfrac{16}{15}$,$\tfrac{19}{10}$},
  ytick={1}, yticklabels={$1$},
  xlabel={inverse depth $q$ on the target ray},
  ylabel={field value at $x(q)$},
  ylabel style={yshift=-4pt},
  title={(b) Exact identified set $\mathcal Q$}]
  \fill[pattern=north east lines, pattern color=gray] (axis cs:0.4,0) rectangle (axis cs:0.8,1.9);
  \fill[black] (axis cs:0.1,-0.04) rectangle (axis cs:0.4,0.04);
  \fill[black] (axis cs:0.8,-0.04) rectangle (axis cs:1.0667,0.04);
  \addplot[gray, thin, domain=0.1:1.9] {x};
  \addplot[black, thick, domain=0.1:1.9, samples=181]
    {min(1.9, min(0.2+2*abs(x-0.5), 0.2+2*abs(x-1.5)))};
  \addplot[black, thick, dashed, domain=0.1:1.9, samples=181]
    {max(0.1, max(0.2-2*abs(x-0.5), 0.2-2*abs(x-1.5)))};
  \addplot[black, densely dotted, thick, domain=0.5:1.3, samples=81]
    {max(0.1, 0.9-2*abs(x-0.9))};
  \addplot[only marks, mark=*, mark size=1.6pt, black] coordinates {(0.9,0.9)};
  \node[font=\scriptsize, anchor=south] at (axis cs:0.6,1.35) {impossible};
  \node[font=\scriptsize, anchor=west] at (axis cs:1.2,1.35) {$q$};
  \node[font=\scriptsize, anchor=south] at (axis cs:1.0,1.22) {$\overline f$};
  \node[font=\scriptsize, anchor=south west] at (axis cs:1.55,0.13) {$\underline f$};
  \node[font=\scriptsize, anchor=west, fill=white, inner sep=1pt] (wit) at (axis cs:1.36,0.97) {$f^{(q)}$};
  \draw[black, thin] (wit.west) -- (axis cs:1.06,0.6);
\end{axis}
\begin{axis}[teaser, name=pc, at={(pb.east)}, anchor=west, xshift=1.55cm,
  xmin=0, xmax=1.45, ymin=0, ymax=3.2,
  xtick={0,1}, xticklabels={$0$,$1$},
  ytick={1,1.5}, yticklabels={$Lh$,$Lh{+}2\varepsilon$},
  xlabel={$\kappa=bL$ (fold parameter)},
  ylabel={sharp $\operatorname{diam}\mathcal Q$},
  ylabel style={yshift=-4pt},
  title={(c) Sharp sampling and noise limits}]
  \fill[pattern=north east lines, pattern color=gray] (axis cs:1,0) rectangle (axis cs:1.45,3.2);
  \addplot[gray, thin] coordinates {(1,0) (1,3.2)};
  \addplot[black, thick, dashed] coordinates {(0,1) (1,1)};
  \addplot[only marks, mark=o, mark size=1.8pt, black] coordinates {(1,1)};
  \addplot[black, thick, domain=0:0.8437, samples=80] {max(1.5, 0.5/(1-x))};
  \node[font=\scriptsize, anchor=south west, align=left] at (axis cs:0.03,1.95) {noisy samples:\\$\max\{Lh{+}2\varepsilon,$\\$\quad 2\varepsilon/(1{-}\kappa)\}$};
  \node[font=\scriptsize, anchor=north west] at (axis cs:0.03,0.95) {exact samples: $Lh$};
  \node[font=\scriptsize, anchor=center, align=center, fill=white, inner sep=1pt] at (axis cs:1.225,2.0) {no uniform\\identification\\as $h\to0$};
\end{axis}
\end{tikzpicture}
\caption{\textbf{Identifying the first visible surface from finite depth samples.}
(a)~Problem: exact depth one at every integer source pixel (dots) is matched
by a smooth surface (solid) whose first hit along a target ray is at depth
$1.659$, while a discrete interval splat returns $[1,1]$ (dashed); pointwise
intervals do not constrain geometry between samples.
(b)~Mechanism: with a slope bound $L$, the lower and upper Lipschitz
envelopes $\underline f,\overline f$ along the ray's epipolar trajectory give
the exact set of attainable first hits,
$\mathcal Q=\{q\ge q_-: q\le\overline f(x(q))\}$ (black bars; example of
\cref{eq:example}). Unlike pointwise bounds or an interval hull, $\mathcal Q$
excludes depths that are necessarily occluded and impossible interior bands
(hatched); each $q\in\mathcal Q$ is realized by an explicit witness scene
$f^{(q)}$ (dotted cone).
(c)~Theory (schematic, not data): on a rectified scanline with spacing $h$
and noise half-width $\varepsilon$, the sharp worst-case diameter is $Lh$
for exact samples up to the fold threshold $\kappa=1$ and
$\max\{Lh+2\varepsilon,2\varepsilon/(1-\kappa)\}$ with noise; above the
threshold, denser sampling cannot remove ambiguity.}
\label{fig:teaser}
\end{figure*}

Depth estimation and view synthesis begin with finite images, yet their
geometric predictions concern continuous surfaces. A pixel value constrains
one source ray. Between measured rays, different surfaces can agree with
every observation while projecting to different depths in a new view.
Visibility makes this uncertainty especially consequential: the desired
quantity is the \emph{first} intersection along a target ray, not an
arbitrary sample-consistent surface point.

This distinction is not repaired by a more sophisticated pointwise
uncertainty model. In a motivating construction, an exact depth of
one at every integer source pixel was consistent with a smooth, one-to-one
reprojection whose true target depth was $1.659$. A discrete interval splat
returned $[1,1]$ (\cref{fig:teaser}a). The construction and its scope appear in
\cref{app:legacy}. It is not evidence that such
aliasing is common; it demonstrates why an explicit between-sample model is
necessary.

We therefore change the inference target. Rather than propose another
confidence score for a nominal warp, we ask for the \emph{identified set}:
every first-hit depth attained by some continuous scene consistent with the
available measurements and a stated regularity bound. This is a
deterministic inverse problem. A learned depth prediction may supply
interval observations, but the theorem applies only when those observations
and the regularity model are valid. An empirical quantile or a measured
finite-difference slope does not establish those premises.

Our main result gives the full set, including its gaps, and a constructive
attainability proof. Parameterizing a target ray by source inverse depth $q$
turns its source-image trajectory into a line $x(q)=e+a q$. If $\lo$ and
$\up$ are the classical lower and upper Lipschitz envelopes and $q_-$ is the
largest fixed point of $\lo(x(q))$, then the exact frontmost inverse-depth
set is
\begin{equation}
\Qcal=\{q\in[q_-,q_{\max}]:q\leq\up(e+a q)\}.
\label{eq:intro-set}
\end{equation}
The lower cutoff enforces visibility. Pointwise lower and upper envelope
tests alone establish only intersection feasibility. A cone anchored at any
$q\in\Qcal$, combined with $\lo$, supplies a complete admissible field
realizing that first hit.

A second result separates source resolution from measurement precision.
Below the folding threshold $bL=1$ in a rectified scanline, exact
measurements identify first-hit inverse depth within $Lh$, independently of
the usual inverse-map amplification factor. Measurement uncertainty,
however, can be amplified by $1/(1-bL)$. We derive the sharp combined bound
rather than multiply both error sources by this factor. At and above the
threshold, explicit indistinguishable surfaces show which convergence claims
fail. We also characterize when a vector of individually feasible target
depths can arise from one shared surface, exposing occlusion constraints
that pointwise intervals cannot express.

The contribution combines exact visibility-sensitive identification,
pointwise and joint witnesses, a finite solver, and sharp sampling/noise
limits (\cref{fig:teaser}). Lipschitz extensions, scalar minimax estimation, and the geometric
possibility of folds are established background. A prespecified
evaluation on ten new scenes and three actual depth estimators separates
solver correctness from failures of the scene assumptions, including
unfavorable results for the fixed learned-depth interval interface.

\section{Related Work}
\label{sec:related}
\paragraph{Optimal recovery from finite information.}
The cone envelopes used here originate in Lipschitz extension
\citep{mcshane1934extension}. Set-membership identification treats bounded
observations and regularity assumptions as a feasible function class
\citep{milanese2004setmembership}. \citet{regier2015miniminimax} give
attainable pointwise envelopes, minimax emulators, and information limits,
while emphasizing that the largest observed slope is only a lower bound on
an unknown Lipschitz constant. We adopt this framework. Our functional is
the first visible intersection after a camera transformation; it can be
discontinuous even when the feasible fields are uniformly close.

\paragraph{First roots and reliable rendering.}
\citet{gal1977root} study optimal sequential and parallel root search when
divided differences have positive lower and upper bounds. In our
subcritical rectified case those bounds are $1-bL$ and $1+bL$. Algorithms
for the first zero of an evaluable function use Lipschitz under- and
overestimators and adaptive interval subdivision
\citep{sergeyev1999twomethods,posypkin2026firstzero}. Sphere tracing uses
regularity to advance safely through a supplied implicit field
\citep{hart1996sphere}. These works preclude a novelty claim for
two-envelope root bounds. Our target is the exact image of a
\emph{fixed-information completion class}, with witnesses for every
attainable first hit and nonconvergence constructions when resolution is
insufficient for visibility.

\paragraph{Geometric uncertainty and reconstruction.}
Space carving characterizes geometry consistent with visual observations
\citep{kutulakos2000spacecarving}; volumetric multi-view stereo enforces
photo-consistency and silhouette constraints to select a reconstruction
\citep{sinha2007adaptive}. Camera-motion robustness certification also
explicitly reasons about z-buffer visibility and single-frame point clouds
\citep{hu2024pixelwise}. Recent visibility-constrained neural reconstruction
fits a Lipschitz implicit field with first-hit supervision
\citep{lenain2026visibility}. These settings differ from our exact
bounded-observation identification problem. In particular, a network's
Lipschitz bound as an implicit field is not a known bound on the
inverse-depth slope of a physical scene.

\paragraph{Sampling and depth prediction.}
Plenoptic sampling and local light-field fusion derive view-sampling rules
for rendering \citep{chai2000plenoptic,mildenhall2019llff}. Sparse-depth
reconstruction exploits representation priors and measurement allocation
\citep{liu2015sparsedepth}. Our sampling variable is spatial resolution of
one source depth sheet, and our loss concerns the first hit in another
view. Modern monocular depth prediction \citep{yang2024dav2} motivates noisy
source observations, but a strong point estimator is not itself a uniform
uncertainty certificate. The learned-depth study tests this interface
empirically without converting marginal calibration into a deterministic
scene guarantee.

\section{Observation and Camera Model}
\label{sec:model}
Let $D\subset\R^d$ be a compact source-image domain, with $d=2$ for a camera
and $d=1$ for a rectified scanline. Fix $0<q_{\min}<q_{\max}<\infty$; the
trivial singleton range is treated in \cref{app:model}. An
inverse-depth field $f:D\to J=[q_{\min},q_{\max}]$ defines a source-visible
radial graph. We observe locations $x_i$ and intervals
$[\ell_i,u_i]\subset J$, and specify $L\geq0$. The feasible class is
\begin{equation}
\Fcal=\{f:D\to J:\Lip(f)\leq L,\quad
                    \ell_i\leq f(x_i)\leq u_i\ \forall i\}.
\label{eq:class}
\end{equation}
Distances are Euclidean in the stated image coordinate system. Intervals
are intersected with $J$ before use; inconsistent observations produce a
refusal. No independence, probability model, or calibration guarantee is
assumed.

\paragraph{Classical extremal fields.}
Define
\begin{align}
\lo(x)&=\max\{q_{\min},\max_i(\ell_i-L\norm{x-x_i})\},\label{eq:lo}\\
\up(x)&=\min\{q_{\max},\min_i(u_i+L\norm{x-x_i})\}.\label{eq:up}
\end{align}
The class is nonempty exactly when $\ell_i-u_j\leq L\norm{x_i-x_j}$ for all
$i,j$, including $i=j$ after clipping to $J$. When nonempty,
$\lo,\up\in\Fcal$ and every $f\in\Fcal$ satisfies $\lo\leq f\leq\up$. These
are classical facts; a proof covering interval observations and degenerate
cases is included for completeness in \cref{app:model}.

\paragraph{Epipolar parameterization.}
Write a target ray in source coordinates as $X_s(z)=c+zv$, where $z>0$ is
target optical-axis depth and the target ray has unit target $z$
coordinate. Assume $v_z>0$ and $1/q_{\max}>c_z$. Then
\begin{align}
z(q)&=\frac{1/q-c_z}{v_z}>0,\label{eq:depth-map}\\
x(q)&=e+a q,\quad e=\frac{v_{xy}}{v_z},\quad
a=c_{xy}-c_z e.\label{eq:epipolar}
\end{align}
These are normalized image coordinates. If source intrinsics give
$x_{\rm pix}=Ax+t_0$, use $e_{\rm pix}=Ae+t_0$ and $a_{\rm pix}=Aa$, with $L$
in pixel units. For a one-dimensional scanline, use the corresponding scalar
coordinates. We require $x(J)\subseteq D$. The scene model is exactly the
radial graph specified by $f$ on $D$, with no additional unmodeled
foreground surface. Since all field values lie in $J$, this line segment
contains every possible intersection with that graph. A failed positivity or
domain check is a refusal, not a certificate of empty space outside the
source view.

An intersection satisfies $f(x(q))=q$. Continuity and the global range imply
at least one intersection: $f(x(q_{\min}))\geq q_{\min}$ and
$f(x(q_{\max}))\leq q_{\max}$. Because $z(q)$ decreases strictly, the
visible hit is
\begin{equation}
q_f=\max\{q\in J:f(e+a q)=q\},\qquad z_f=z(q_f).
\label{eq:firsthit}
\end{equation}
Tangencies and intervals of roots are included. The identified sets are
$\Qcal=\{q_f:f\in\Fcal\}$ and $\Zcal=z(\Qcal)$. These are pointwise sets for
a target ray; arbitrary independently chosen values at multiple rays need
not come from the same scene.

\section{Exact First-Hit Identification}
\label{sec:identification}
Let $q_-$ and $q_+$ denote the greatest roots of $\lo(e+a q)=q$ and
$\up(e+a q)=q$, respectively. The extrema of the class supply sharp
endpoint bounds, but the interior of their interval need not be attainable.

\begin{theorem}[Full identified set and witnesses]
\label{thm:fullset}
Under \cref{sec:model}, with $\Fcal\ne\varnothing$,
\begin{equation}
\Qcal=[q_-,q_{\max}]\cap\{q:q\leq\up(e+a q)\}.
\label{eq:fullset}
\end{equation}
Every $q\in\Qcal$ is the first hit of the globally admissible field
\begin{equation}
f^{(q)}(x)=\max\{\lo(x),q-L\norm{x-(e+a q)}\}.
\label{eq:witness}
\end{equation}
The minimum and maximum of $\Qcal$ are $q_-$ and $q_+$. Its sharp
target-depth hull is $[z(q_+),z(q_-)]$.
\end{theorem}
\begin{proof}[Proof idea]
After its last root, $\lo(e+a t)<t$. Every admissible field lies above
$\lo$, so continuity forces a root at least as large as $q_-$. Any root must
also lie below $\up$ at its source projection. Conversely, the cone in
\cref{eq:witness} is below $\up$ everywhere because $\up$ is $L$-Lipschitz
and $q\leq\up(e+a q)$. Its maximum with $\lo$ therefore satisfies all
observations and the global range. It equals $q$ at the anchor. For every
$t>q$, both terms are strictly below $t$, so no closer target intersection
is possible. This proves attainability, including when the scene projection
folds. The complete proof is in \cref{app:identification}.
\end{proof}

\paragraph{Why the full set matters.}
The condition $\lo(e+a q)\leq q\leq\up(e+a q)$ identifies feasible
\emph{intersections}; it can include depths that are necessarily occluded.
The cutoff $q\geq q_-$ removes them. In addition, the upper-envelope
condition can remove interior bands. For example, take $J=D=[1/10,19/10]$,
$e=0$, $a=1$, $L=2$, and exact observations $f(1/2)=f(3/2)=1/5$. Then
\begin{equation}
\Qcal=[1/10,2/5]\ \cup\ [4/5,16/15].
\label{eq:example}
\end{equation}
For $z=1/q$, the depth interval $(5/4,5/2)$ is impossible, despite lying
inside the sharp interval hull. The formula does not assign probabilities to
the two components.

\begin{proposition}[Connectedness and output size]
\label{prop:connected}
If $\kappa=L\norm{a}\leq1$, then $\Qcal=[q_-,q_+]$; if $\kappa<1$, every
admissible field has a unique intersection. In general, $n$ observations
yield at most $n+1$ connected components. For every $\kappa>1$, this
component bound is attainable even with exact observations.
\end{proposition}
The first claim follows because $f(e+a q)-q$ is nonincreasing when
$\kappa\leq1$. For the component bound, each excluded cone set
$\{q:q>u_i+L\norm{e+a q-x_i}\}$ is an open interval. Removing at most $n$
such intervals gives the bound. Sharpness is proved in
\cref{app:solver}.

\subsection{Several Rays Must Share a Scene}
For rays $x_j(q)=e_j+a_jq$ satisfying \cref{sec:model} and proposed inverse
depths $\widetilde q_j\in J$, append exact observations
$(e_j+a_j\widetilde q_j,\widetilde q_j)$ to the original information. Let
$\Fcal_{\rm aug}$ and $\lo_{\rm aug}$ be its feasible class and lower
envelope. Compatible anchor values are necessary, but visibility adds a
further condition.

\begin{theorem}[Exact joint first-hit membership]
\label{thm:joint}
The proposed vector is attainable as the first-hit vector of one field in
$\Fcal$ if and only if $\Fcal_{\rm aug}\ne\varnothing$ and
\begin{equation}
\max\{q\in J:\lo_{\rm aug}(e_j+a_jq)=q\}=\widetilde q_j
\quad\text{for every }j.
\label{eq:joint}
\end{equation}
When these conditions hold, $\lo_{\rm aug}$ itself is a joint witness.
\end{theorem}
Any realizing field must satisfy the anchors and lie above $\lo_{\rm aug}$.
The lower envelope's greatest root is therefore no larger than the
realizing field's first-hit inverse depth, while the anchor supplies a root
at $\widetilde q_j$. This proves necessity; rendering $\lo_{\rm aug}$ proves
sufficiency. A full proof and same-camera counterexample are in
\cref{app:joint}. In that example, two individually
possible and Lipschitz-compatible anchors force an earlier occluder on the
first ray: the proposed inverse depths $(1/4,1)$ cannot occur jointly,
whereas $(5/8,1)$ can. Thus treating pointwise supports as an unconstrained
Cartesian product loses geometric information. The two endpoint maps are a
special exception: $\lo$ and $\up$ attain all far and near endpoints
simultaneously for a common scene class and fixed pose.

\paragraph{Optimal prediction and abstention.}
Let $d_-=z(q_+)$ and $d_+=z(q_-)$. Since both endpoints are attained, the
exact minimax relative-depth estimate and error are
\begin{equation}
\widehat d^*=\frac{2d_-d_+}{d_-+d_+},\qquad
R^*=\frac{d_+-d_-}{d_++d_-}.
\label{eq:minimax}
\end{equation}
A relative-error certificate at tolerance $\tau<1$ is possible from this
information if and only if $d_+/d_-\leq(1+\tau)/(1-\tau)$. Otherwise every
deterministic estimator fails on an admissible completion. This elementary
minimax consequence is not a separate novelty claim. Gaps in $\Zcal$ matter
for set-valued or depth-band decisions; they do not improve this scalar
worst-case loss over the sharp hull. \Cref{app:decisions}
proves these assertions.

\begin{figure*}[t]
\centering
\includegraphics[width=\linewidth]{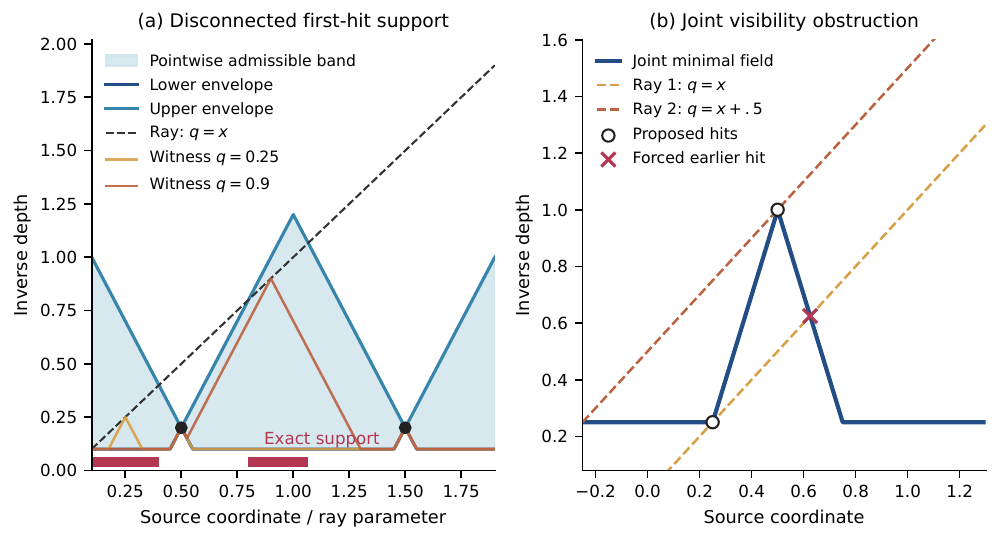}
\caption{\textbf{Exact support and shared visibility.} Left: two exact
observations leave a disconnected first-hit support; an interval hull
includes an impossible band. Right: individually possible, mutually
Lipschitz-compatible depths can still be impossible as joint first hits.
The forced lower cone creates an earlier occluder. These are analytic
constructions, not measured prevalence.}
\label{fig:support}
\end{figure*}

\section{Sharp Sampling and Noise Limits}
\label{sec:resolution}
Consider a rectified scanline with projection $F_f(x)=x+b f(x)$, $b>0$, and
$\kappa=bL$. Observations at $x_i=A+ih$ have common half-width $\eps$ around
arbitrary centers $y_i$. The feasible class is nonempty, and every
considered field intersects the target coordinate $v$ inside the sampled
domain. Then its first-hit inverse depth is $(v-x_f)/b$, where $x_f$ is the
smallest root of $F_f(x)=v$.

\begin{theorem}[Sharp subcritical uncertainty]
\label{thm:grid}
For $\kappa<1$, the first-hit identified-set diameter obeys
\begin{equation}
\diam(\Qcal_v)\leq
\max\left\{Lh+2\eps,\frac{2\eps}{1-\kappa}\right\}.
\label{eq:grid}
\end{equation}
Both branches are attained by admissible fields when the source domain and
positive depth range accommodate the extremizers. For exact samples, the
sharp bound is $Lh$ for all $\kappa\leq1$; at equality it is a supremum and
need not be attained.
\end{theorem}
\begin{proof}[Main argument]
For two compatible fields with roots $x_f\leq x_g$, write $w=(x_g-x_f)/b$.
If a grid node $s$ lies between them, Lipschitz continuity gives
$f(s)-g(s)\geq(1-\kappa)w$. The common sample interval has width $2\eps$,
giving the second branch. Otherwise both roots lie in one cell $[a,a+h]$.
Its two endpoint observations give
\[
w\leq2\eps+L\min\{x_f+x_g-2a,2a+2h-x_f-x_g\},
\]
which is at most $2\eps+Lh$. For $\kappa=1$ and exact samples, monotonicity
and the common grid values still put all first roots in the same cell.
Sharpness is proved in \cref{app:sampling}.
\end{proof}

\paragraph{Resolution and precision are different resources.}
For $L>0$, the branches exchange dominance at $2b\eps=(1-\kappa)h$; when
$L=0$ both equal $2\eps$. Multiplying the entire interpolation error by
$1/(1-\kappa)$ would be unnecessarily loose: with exact samples, the sharp
diameter stays $Lh$ as $\kappa\uparrow1$. For noisy samples, increased
spatial density cannot remove the worst-case precision floor
$2\eps/(1-\kappa)$. These are fixed-grid limits; no claim of optimality over
all adaptive sensing policies is made.

\begin{theorem}[Loss of uniform identification]
\label{thm:obstruction}
For every $bL>1$, a fixed source domain, positive range, and target admit
pairs of $L$-Lipschitz fields agreeing on uniform grids with $h\to0$ whose
first-hit inverse depths differ by a quantity tending to $1/(2b)$. At
$bL=1$, arbitrarily small positive interval noise can also leave a
nonvanishing first-hit separation.
\end{theorem}
\Cref{app:sampling} supplies explicit
piecewise-linear fields and complete proofs. In the supercritical example, a
bump vanishing at every sample creates an earlier intersection while a later
one persists. The corresponding two-point minimax relative-depth error tends
to $1/7$. Thus dense sampling alone cannot provide uniform worst-case
consistency above the threshold. The adversarial fields depend on $h$; this
is not a claim that a fixed well-separated scene fails to converge. Known
regularity and a crossing margin can still make individual queries
informative.

\begin{figure*}[t]
\centering
\includegraphics[width=\linewidth]{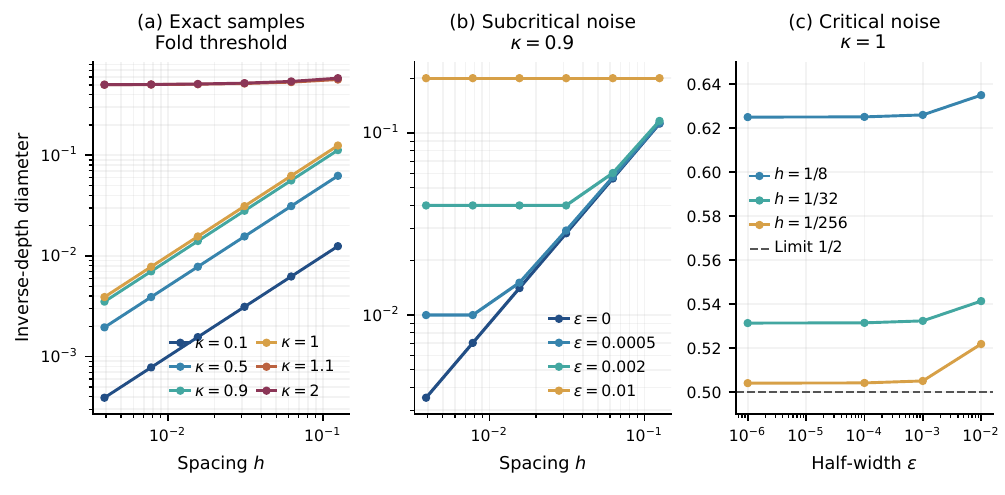}
\caption{\textbf{Resolution, noise, and folds.} Analytic extremizers and
computed identified sets show the two sharp subcritical regimes, critical
sensitivity to positive noise, and persistent ambiguity above the fold
threshold. Markers are computed support diameters for the proved analytic
families, with lines connecting them; dashed limits are labeled.
Domain/range clipping is avoided in sharpness cases.}
\label{fig:phase}
\end{figure*}

\section{Computing the Identified Set}
\label{sec:algorithm}
For each sample, form
\begin{align}
A_i&=\{q\in J:q+L\norm{e+a q-x_i}\leq\ell_i\},\label{eq:A}\\
B_i&=\{q\in J:q-u_i-L\norm{e+a q-x_i}>0\}.\label{eq:B}
\end{align}
Each $A_i$ is a closed interval and each $B_i$ a relatively open interval,
possibly empty. Their boundaries solve scalar quadratic equations. Squaring
is valid only after retaining $q\leq\ell_i$ for $A_i$ and $q>u_i$ for $B_i$.
Linear, constant, and tangent cases are treated explicitly.

\begin{figure}[t]
\refstepcounter{algorithm}
\label{alg:identify}
\hrule\smallskip
\textbf{Algorithm \thealgorithm. First-hit identification}
\par\smallskip
\textbf{Input:} Feasible samples, $L,J,D$; target ray $c,v$.
\begin{enumerate}[leftmargin=1.4em,itemsep=2pt,topsep=3pt]
\item Check $v_z>0$, $1/q_{\max}>c_z$, and $e+aJ\subset D$.
If any fails, return \textsc{OutsideModel}.
\item Set $r\gets q_{\min}$ and $\mathcal B\gets\varnothing$.
\item For each sample $i$, compute $A_i,B_i$ with unsquared cone
inequalities; set $r\gets\max(r,\max A_i)$ if $A_i\ne\varnothing$,
and add nonempty $B_i$ to $\mathcal B$.
\item Set $\Qcal\gets[r,q_{\max}]\setminus\bigcup_{B\in\mathcal B}B$.
\item Return $z(\Qcal)$, \cref{eq:minimax}, and \cref{eq:witness}.
\end{enumerate}
\smallskip\hrule
\end{figure}

\Cref{app:solver} proves the algorithm and its
$O(nd+n\log n)$ arithmetic cost per admissible ray after data feasibility
has been checked. A direct pairwise feasibility check costs $O(n^2d)$ once
per source information set. The segment-domain check takes $O(d)$ endpoint
tests for the interval/rectangle domains used here; for a general compact
domain it is a separate required check. The algorithm's output can itself
have $n+1$ components. For one-dimensional data the cone constraints reduce
to affine inequalities; the accompanying independent oracle computes their
endpoints exactly as rational numbers. For $m$ proposed target depths,
\cref{thm:joint} checks the additional anchor constraints and $m$
lower-envelope roots in $O((mn+m^2)d)$ arithmetic operations after the
original feasibility check. It does not search over all possible depth
vectors.

\paragraph{Numerical and model refusals.}
The empirical implementation uses double precision and a documented small
outward tolerance for near-coincident endpoints. It is a numerical
reference, not a directed-rounding machine certificate. It preserves
tangent and singleton components; open excluded intervals that merely touch
cannot be merged as if their common endpoint were impossible. Infeasible
intervals, an unobserved ray segment, nonpositive candidate depths, and an
invalid pose orientation have separate return conditions. No fallback
silently imputes an unseen foreground or replaces an inconsistent class with
a chosen interpolant. The method can be applied to an explicitly declared
smooth patch, but patch selection and the absence of earlier outside-patch
geometry are additional assumptions, measured separately in the empirical
protocol.

\section{Computational Evidence}
\label{sec:experiments}
We separate arithmetic verification, analytic stress cases, and
natural-scene diagnostics. No experiment converts a chosen regularity bound
or empirical source band into a physical-scene guarantee. Complete
protocols, denominators, and results are in
\cref{app:protocol,app:extended-results,app:reexecution}.

\begin{table*}[t]
\centering\small
\begin{tabular}{lrrlrr}
\toprule
Source observations & Queries & In class (\%) & Accept at 5\% (\%) & Digital in-class & Measured $>$5\% errors \\
\midrule
Reference bands & 14,700 & 22.97 & 30.78 [19.44, 44.33] & 4,170/4,170 & 0/5,380 \\
StereoSGBM & 4,900 & 4.05 & 33.82 [21.29, 46.69] & 265/265 & 95/1,986 \\
Depth Anything V2 & 4,900 & 3.87 & 2.82 [0.80, 5.86] & 225/225 & 169/169 \\
ZoeDepth & 4,900 & 4.61 & 9.92 [5.12, 15.77] & 295/295 & 409/503 \\
\bottomrule
\end{tabular}
\caption{\textbf{Completed ten-scene primary evaluation.} In-class and
acceptance rates are equal-scene means; brackets are descriptive 95\%
scene-bootstrap intervals. Coverage and error columns are pooled counts with
explicit denominators. Measured error uses accepted queries with an
available right-view reference. Reference bands (E1a) have three noise
replicates; the other studies have one. Digital membership is an oracle evaluation label, not a
selection rule. Ten SGBM in-class records have no supplied observations. All
primary cells have $L\beta=1$, hence connected support. These empirical
rates do not establish a physical-scene certificate.}
\label{tab:heldout-primary}
\end{table*}

\paragraph{Independent arithmetic and witness checks.}
An exact rational one-dimensional oracle uses affine half-space
intersection and rational endpoints, independently of the numerical solver.
Across 600 seeded, consistent rational cases, including zero and critical
slopes and supports with up to eight components, all supports agreed; the
largest numerical endpoint difference was $3.47\times10^{-14}$. In a separate 800-case suite (500 scanlines, 300
two-dimensional source domains), 2,064 constructed witnesses had no
constraint failures, with maximum floating residual $1.13\times10^{-13}$.
These are implementation checks, not proofs.

\paragraph{Sharpness and joint consistency.}
The solver reproduced 72 subcritical sharp-bound fixtures and 30 exact
critical fixtures to absolute error below $8.80\times10^{-14}$. Twenty-four
noisy critical and 12 exact supercritical cases retained the prescribed
separated first-hit witnesses (\cref{fig:phase}). For 2,000 individually
supported depth pairs in the fixed folded example, 311 failed anchor
consistency and a further 308 failed only the visibility condition; 1,381
were jointly attainable. All 300 independently constructed jointly feasible
scenes passed. These counts illustrate the distinction between constraints;
their sampling distribution is artificial.

\paragraph{Development diagnostic.}
A three-scene Middlebury study \citep{scharstein2014middlebury} used
reference-derived bands and explicitly piecewise-linear digital scenes. Only
425 of 2,100 primary queries (20.2\%) had complete reference membership in
the assumed class; all were contained. Mean minimax point error was
0.322\%, versus 0.0847\% for linear interpolation on those same queries. Its
full 81-cell record includes 125 out-of-class misses and no in-class misses.
\Cref{app:extended-results} retains every setting and
refusal. Conditional correctness, restricted applicability, and the absence
of an average point-accuracy gain are all preserved.

\paragraph{Ten-scene estimator and task validation.}
The held-out design was fixed before acquiring the new Middlebury scenes
(\cref{app:protocol}). It uses metadata-only window sizing and
integer target rays $x(q)=e+\beta q$, where $\beta\in\{1/4,1/2,1\}$ is the
fraction of the stereo baseline used for the virtual target view ($\beta=1$
is the measured right view), so $\kappa=L\beta$. Study E1a supplies
reference-derived bands with bounded noise and E1b fixed StereoSGBM
predictions \citep{opencv2024sgbm}; study E2 supplies fixed relative bands
around two pinned metric monocular models, Depth Anything V2 and
ZoeDepth \citep{yang2024dav2,bhat2023zoedepth}. No parameters, scale
alignment, or source bands are chosen from target error.
\Cref{tab:heldout-primary} retains every planned query. Equal-scene mean
class membership is 22.97\% for reference bands and only 4.05\%, 3.87\%, and
4.61\% for the three real-estimator interfaces. All in-class digital first
hits are contained. Measured right-view truth is a separate diagnostic: at
the fixed five-percent tolerance, SGBM has 95 violations among 1,986
accepted evaluable queries. The learned interfaces have 169/169 and 409/503
violations. Their native units and postprocessing were audited; these
unfavorable results are preserved. They demonstrate that narrow conditional
sets can be incorrect when source bands or scene assumptions fail, not a
failure of the conditional theorem.

\paragraph{Visibility structure appears in the predicted regime.}
The primary cell has $L\beta=1$: full support is connected, as the theory
predicts, and its observed first-hit/all-intersection hull comparison is
null. The latter equality is not guaranteed at the critical value: root
plateaus can separate the two sets (\cref{app:solver}). In the
prespecified sensitivity grid, changing only $L$ to four gives equal-scene
disconnected-output rates of 92.06\% for reference bands and 59.99\% for
SGBM. The mean of scene-median impossible-gap fractions inside the depth
hull is 23.58\% and 13.92\%; corresponding median depth-hull reductions
against all intersections are 33.73\% and 16.91\%. These are descriptive
summaries over feasible outputs, with scene intervals and all settings in
\cref{app:extended-results}. They establish a material
visibility effect inside the specified completion classes. They do not
license replacing the frozen primary result or interpreting invalid source
bands as physical uncertainty bounds.

\paragraph{Independent re-execution and optional studies.}
An independent re-execution (fp32, batch size one)
reproduced every one of the \EtwoRecords{} frozen E2 records per learned
model: all membership, feasibility, refusal and acceptance decisions are
identical, and continuous outputs differ by at most \EtwoMaxRel{} relative.
The optional general-pose study O1 contained all \OoneMembers{} evaluable
digital class members among \OoneQueries{} planned queries, with no accepted
error above five percent against the digital truth. The pose-stress study
O2 refused every query under its executable two-dimensional form and is
uninformative; a labelled post-hoc wide-window variant (O2-w) is reported
separately as exploratory. \Cref{tab:optional} and \cref{app:reexecution}
report these outcomes and extensions to
Depth Pro \citep{bochkovskii2025depthpro} and TUM RGB-D \citep{sturm2012tum}.

\paragraph{Motivating construction.}
On the smooth between-sample construction of \cref{sec:intro}
(\cref{app:legacy}), the solver contains the true depth but
refuses a 25\% relative-error tolerance: its minimax uncertainty is
30.42\%. The information is insufficient, and the output says so.

\section{Limitations and Conclusions}
\label{sec:limitations}
\paragraph{Limitations.}
All guarantees are conditional on the specified scene class: a known pose,
a global slope bound, valid interval observations, a positive depth range,
and a single source-visible sheet with no unmodeled foreground. Source
consistency cannot establish these premises between samples, and the
evaluation shows that fixed learned-depth bands often violate them,
yielding narrow but incorrect sets. The sharp resolution/noise theorem is
specific to a rectified scanline.

\paragraph{Conclusions.}
Finite measurements can determine more than a nominal interpolation and
less than a narrow interval certificate suggests. The exact first-hit set,
its globally consistent witnesses, and the sharp distinction between spatial
resolution and measurement precision identify that boundary. They also state
when a tighter answer is impossible from the same information.

\section*{LLM Usage Statement}
The authors developed the research ideas and proofs. AI tools provided coding, debugging, writing, and editing support, including polishing the exposition and presentation of the proofs. The authors reviewed and verified the resulting material and retain responsibility for the research questions, methods, proofs, experiments, interpretation, and final content.

{
    \small
    \bibliographystyle{ieeenat_fullname}
    \bibliography{references}
}

\clearpage
\onecolumn
\appendix
\counterwithin{theorem}{section}
\counterwithin{equation}{section}
\counterwithin{table}{section}
\counterwithin{figure}{section}

\section*{Appendices}
\Cref{app:model,app:identification,app:solver,app:joint,app:decisions,app:sampling,app:scope}
contain the complete proofs and constructions. \Cref{app:protocol} gives
the prespecified validation protocol, \cref{app:extended-results} the complete
computational results and sensitivity grids, \cref{app:reexecution} the
independent re-execution and optional studies, \cref{app:legacy} the
motivating aliasing construction, and \cref{app:reproducibility}
reproducibility and asset provenance.

\section{Model, information, and exact source envelopes}
\label{app:model}

Let $D\subset\mathbb R^d$ be a source-image domain, with $d=2$ for an
image and $d=1$ for a scanline. Distances below are Euclidean in the
specified image coordinate system. Fix $L\geq0$ and a positive inverse-depth
range $Q=[q_{\min},q_{\max}]$, where $0<q_{\min}<q_{\max}<\infty$.
We observe locations $x_1,\ldots,x_n\in D$ and intervals
$[\ell_i,u_i]\subseteq Q$. The admissible inverse-depth fields are
\begin{equation}
\mathcal F=\{f:D\to Q:\ |f(x)-f(y)|\leq L\|x-y\|\ \text{for all }x,y\in D,
\quad \ell_i\leq f(x_i)\leq u_i\ \text{for all }i\}.
\label{eq:app-class}
\end{equation}
Intervals may be obtained by intersecting original intervals with $Q$;
an empty intersection makes the information inconsistent. The following
classical envelopes are included with proofs to specify exactly the
information used. Their construction is not a novelty claim.
Define
\begin{align}
f_-(x)&=\max\{q_{\min},\max_i[\ell_i-L\|x-x_i\|]\},\label{eq:app-lower}\\
f_+(x)&=\min\{q_{\max},\min_i[u_i+L\|x-x_i\|]\}.\label{eq:app-upper}
\end{align}

\begin{lemma}[Feasibility and extremal fields]
\label{lem:app-envelopes}
The class $\mathcal F$ is nonempty if and only if
\begin{equation}
\ell_i-u_j\leq L\|x_i-x_j\|\qquad\text{for every ordered pair }(i,j).
\label{eq:app-feasibility}
\end{equation}
When this condition holds, $f_-,f_+\in\mathcal F$, and every
$f\in\mathcal F$ satisfies $f_-\leq f\leq f_+$ pointwise. Thus both
pointwise envelopes are attained by globally admissible fields.
\end{lemma}
\begin{proof}
If $f\in\mathcal F$, then
$\ell_i\leq f(x_i)\leq f(x_j)+L\|x_i-x_j\|\leq u_j+L\|x_i-x_j\|$,
which proves necessity. The same Lipschitz inequalities give
$\ell_i-L\|x-x_i\|\leq f(x)\leq u_i+L\|x-x_i\|$,
and hence the pointwise bounds.
For sufficiency, a finite maximum or minimum of functions with the same
Lipschitz bound $L$ is $L$-Lipschitz. Moreover $f_-\geq q_{\min}$,
$f_-\leq q_{\max}$, and $f_-(x_i)\geq\ell_i$. Condition
\eqref{eq:app-feasibility} implies
$\ell_j-L\|x_i-x_j\|\leq u_i$ for every $j$, so
$f_-(x_i)\leq u_i$. Thus $f_-\in\mathcal F$.
The symmetric argument proves $f_+\in\mathcal F$.
The pointwise bounds already proved then imply $f_-\leq f_+$.
\end{proof}

\paragraph{Singleton range.}
If $q_{\min}=q_{\max}=q_0>0$, feasible clipped observations all equal
$q_0$ and the class contains only the constant field $f\equiv q_0$.
For an admissible ray its identified set is the singleton $\{q_0\}$,
with zero minimax error. The nondegenerate statements below assume a
strict range interval to avoid repeatedly separating this trivial case.

\paragraph{Scope of the scene model.}
In normalized image coordinates a field $f$ represents the source-visible
radial surface
\begin{equation}
\mathcal S_f=\{(x_1/f(x),x_2/f(x),1/f(x)):x\in D\}.
\label{eq:app-surface}
\end{equation}
Our first-hit statements concern intersections with this surface model.
Self-occlusion in the target view is allowed. Unobserved additional objects,
back surfaces, and geometry outside $D$ are not identified by the observations.
Interpreting a model certificate as a certificate for the entire physical
scene requires independently excluding any such earlier occluder along
the target ray. No finite-sample theorem below removes that requirement.
Likewise $L$ and $Q$ are model assumptions, not upper bounds justified by
taking the maximum adjacent slope or the observed range of a sampled image.
A test of consistency with the observations can reject an assumption;
it cannot verify it between observations.

\section{Calibrated rays and the exact identified set}
\label{app:identification}

Suppose a relative pose maps source coordinates to target coordinates by
$X_t=RX_s+t$, and let $r$ be a target ray with target-camera $z$ component
one. Write
\begin{equation}
v=R^\top r,\qquad c=-R^\top t.
\end{equation}
A point at target optical-axis depth $z$ has source coordinates
$X_s=zv+c$. Assume
\begin{equation}
v_z>0,\qquad q_{\max}^{-1}-c_z>0.
\label{eq:app-positive}
\end{equation}
Parameterize the ray by inverse source depth $q=1/(zv_z+c_z)$.
Its target depth and source image position are respectively
\begin{align}
z(q)&=\frac{q^{-1}-c_z}{v_z},\label{eq:app-zq}\\
x(q)&=e+aq,\quad
 e=\frac{v_{xy}}{v_z},\quad a=c_{xy}-c_z\frac{v_{xy}}{v_z}.
\label{eq:app-epipolar}
\end{align}
If the source camera's affine intrinsics map is $x_{\rm pix}=Ax+t_0$,
then $e_{\rm pix}=Ae+t_0$ and $a_{\rm pix}=Aa$. In particular the
principal-point translation does not act on the direction $a$.
The Lipschitz bound $L$ must be specified in those same coordinates.
Assume the entire epipolar segment is observed:
\begin{equation}
x(Q)\subseteq D.\label{eq:app-domain}
\end{equation}
The conditions are checked for each query. The results stated here do not
cover $v_z\leq0$, a nonpositive target-depth candidate, or an epipolar
segment leaving the modeled source domain.

\begin{lemma}[Ray reduction and existence]
\label{lem:app-ray}
Under \eqref{eq:app-positive}--\eqref{eq:app-domain}, intersections of the
target ray with $\mathcal S_f$ are in one-to-one correspondence with the
roots of $f(e+aq)=q$ in $Q$. There is at least one such root for every
$f\in\mathcal F$. Its frontmost intersection has inverse source depth
\begin{equation}
q_f=\max\{q\in Q:f(e+aq)=q\}\label{eq:app-front}
\end{equation}
and target depth $z(q_f)$.
\end{lemma}
\begin{proof}
Solving $q^{-1}=zv_z+c_z$ gives \eqref{eq:app-zq}. Substituting into
$(zv_{xy}+c_{xy})/(zv_z+c_z)$ gives \eqref{eq:app-epipolar}.
A point on the ray lies on the radial graph exactly when its inverse
source depth equals the graph value at its source projection. This proves
the correspondence. The continuous function $g_f(q)=f(e+aq)-q$ obeys
$g_f(q_{\min})\geq0$ and $g_f(q_{\max})\leq0$, since $f$ takes values in
$Q$. The intermediate value theorem gives a nonempty root set. It is
closed in the compact interval $Q$, so its maximum exists.
Finally $z(q)>0$ and $z'(q)=-1/(v_zq^2)<0$, so the greatest root is the
nearest positive target-depth intersection with the modeled surface.
\end{proof}

Define the two extremal roots
\begin{equation}
q_- =\max\{q\in Q:f_-(e+aq)=q\},\qquad
q_+ =\max\{q\in Q:f_+(e+aq)=q\}.\label{eq:app-endpoints}
\end{equation}
These are last roots in $q$, even when the projected surface folds.

\begin{theorem}[Exact first-hit identification, including folds]
\label{thm:app-fullset}
Assume $\mathcal F\ne\varnothing$ and the geometric conditions above.
The set of all frontmost inverse source depths compatible with the
observations is exactly
\begin{equation}
\{q_f:f\in\mathcal F\}
=\mathcal Q
=\{q\in[q_-,q_{\max}]:q\leq f_+(e+aq)\}.\label{eq:app-fullset}
\end{equation}
Every point $q\in\mathcal Q$ has the explicit globally admissible witness
\begin{equation}
f^{(q)}(x)=\max\{f_-(x),q-L\|x-(e+aq)\|\}.\label{eq:app-witness}
\end{equation}
The minimum and maximum of $\mathcal Q$ are $q_-$ and $q_+$, respectively.
Consequently the sharp target-depth identified set is $z(\mathcal Q)$,
and its sharp interval hull is $[z(q_+),z(q_-)]$.
\end{theorem}
\begin{proof}
First observe that
\begin{equation}
f_-(e+at)<t\qquad(q_-<t\leq q_{\max}).\label{eq:app-afterlast}
\end{equation}
Equality would contradict the definition of $q_-$. If the left side
were greater at some such $t$, continuity and
$f_-(e+aq_{\max})\leq q_{\max}$ would give a still later root, again a
contradiction. This proves \eqref{eq:app-afterlast}.

For any $f\in\mathcal F$, at $q_-$ we have
$f(e+aq_-)\geq f_-(e+aq_-)=q_-$. At $q_{\max}$ its value is at most
$q_{\max}$. Thus $f$ has a root at or above $q_-$, and its greatest root
$q_f$ is at least $q_-$. At that root
$q_f=f(e+aq_f)\leq f_+(e+aq_f)$. This proves inclusion in the right
side of \eqref{eq:app-fullset}.

Conversely, take a $q$ in that right side and put $x_q=e+aq$.
By \eqref{eq:app-afterlast}, including equality at $q=q_-$, we have
$f_-(x_q)\leq q$. Also $q\leq f_+(x_q)$.
The function in \eqref{eq:app-witness} is $L$-Lipschitz. It is at least
$f_-$ and at most $f_+$: indeed the Lipschitz property of $f_+$ gives
\[
q-L\|x-x_q\|\leq f_+(x_q)-L\|x-x_q\|\leq f_+(x).
\]
It therefore lies in $Q$ and satisfies every sample interval, hence is
in $\mathcal F$. At $x_q$ it equals $q$. For every $t>q$,
\[
f_-(e+at)<t,\qquad
q-L\|(e+at)-(e+aq)\|=q-L\|a\|(t-q)<t.
\]
Both terms in its maximum are strictly below $t$, so this witness has no
root larger than $q$. Its frontmost root is exactly $q$.

The point $q_-$ is in $\mathcal Q$, since $f_+\geq f_-$.
The maximum of the set where $f_+(e+aq)-q\geq0$ is its greatest root:
a strictly positive value at its maximum would, by continuity and the
nonpositive right-endpoint value, force a larger member. Thus the
maximum is $q_+$. Monotonicity of $z$ proves the depth conclusions.
\end{proof}

\begin{remark}[Information, rather than a chosen interpolant]
The witness establishes necessity as well as sufficiency. A depth excluded
by \eqref{eq:app-fullset} is impossible under the assumptions; a depth
included in it occurs for a whole admissible source field sharing all
observed intervals. Rendering one fixed interpolant, even exactly, does
not provide this identification statement. Conversely, the result does
not say that all completions are equally probable: no distribution over
completions is assumed.
\end{remark}

\section{Exact scalar solver and connected components}
\label{app:solver}

Put $b_i=e-x_i$, $\kappa=L\|a\|$, and define
\begin{align}
A_i&=\{q\in Q:q+L\|aq+b_i\|\leq\ell_i\},\label{eq:app-Ai}\\
B_i&=\{q\in Q:q-u_i-L\|aq+b_i\|>0\}.\label{eq:app-Bi}
\end{align}
Empty $A_i$ contribute nothing to a maximum. Then
\begin{equation}
q_-=\max\bigl(\{q_{\min}\}\cup\bigcup_i A_i\bigr),\qquad
\mathcal Q=[q_-,q_{\max}]\setminus\bigcup_i B_i.\label{eq:app-solver}
\end{equation}

\begin{proposition}[Finite exact computation]
\label{prop:app-algorithm}
Each $A_i$ is a closed interval or empty, and each $B_i$ is an open
interval relative to $Q$ or empty. Their endpoints can be obtained from
scalar quadratic equations, with the unsquared sign conditions retained.
After data feasibility and geometric admissibility have been checked,
\eqref{eq:app-solver} takes
$O(nd+n\log n)$ arithmetic operations per ray in the real-arithmetic
model. Its output has at most $n+1$ connected components. In particular,
it does not require ray marching, a depth grid, or selecting one root
by a local solver.
\end{proposition}
\begin{proof}
The function $q+L\|aq+b_i\|$ is convex, so its sublevel set $A_i$ is
convex and closed in the compact interval $Q$. The function
$q-u_i-L\|aq+b_i\|$ is concave, so its strict superlevel set $B_i$ is
convex and relatively open. A convex subset of a line is an interval.
For $q>q_{\min}$, the condition $f_-(e+aq)\geq q$ is equivalent to
membership in at least one $A_i$; $q=q_{\min}$ is always included.
The rightmost point of this nonnegative set is the last root, proving
the first formula in \eqref{eq:app-solver}. The second formula is
\eqref{eq:app-fullset} expanded using $f_+$.

For $A_i$ retain $q\leq\ell_i$ and solve
\begin{equation}
(\kappa^2-1)q^2+2(L^2a^\top b_i+\ell_i)q
 +L^2\|b_i\|^2-\ell_i^2\leq0.\label{eq:app-lowerquad}
\end{equation}
These two conditions are equivalent to \eqref{eq:app-Ai}.
For $B_i$ retain $q>u_i$ and solve
\begin{equation}
(1-\kappa^2)q^2-2(u_i+L^2a^\top b_i)q
 +u_i^2-L^2\|b_i\|^2>0.\label{eq:app-upperquad}
\end{equation}
These are equivalent to \eqref{eq:app-Bi}. Degenerate linear or constant
polynomials are handled as such. There are $O(n)$ endpoints; sorting
and taking the complement of the open intervals costs $O(n\log n)$.
Removing $n$ intervals from one interval leaves at most $n+1$ components.
\end{proof}

\paragraph{Endpoint and numerical conventions.}
The removed sets are open. A tangent root or a singleton between two
open excluded intervals remains admissible. Therefore an implementation
must retain endpoint membership and must not silently merge intervals
that merely touch. A convenient implementation obtains polynomial roots
and sign-change candidates, then checks each candidate point and open
cell in the original unsquared inequalities. The complexity statement
concerns exact real arithmetic. Standard floating-point execution and
numerical residual tests are evidence about an implementation; without
directed rounding or interval predicates they are not formal machine
certificates.

\begin{proposition}[A sharp connectivity threshold]
\label{prop:app-connectivity}
If $\kappa\leq1$, then $\mathcal Q=[q_-,q_+]$. If $\kappa<1$, each
admissible field has exactly one intersection on the considered ray.
For every $\kappa>1$ and every positive integer $n$, there exist
consistent exact samples for which $\mathcal Q$ has $n+1$ components.
\end{proposition}
\begin{proof}
For $t>s$ and any $f\in\mathcal F$,
\[
[f(e+at)-t]-[f(e+as)-s]\leq (\kappa-1)(t-s).
\]
Thus the residual is nonincreasing for $\kappa\leq1$, and strictly
decreasing for $\kappa<1$. Existence was established in
Lemma~\ref{lem:app-ray}. Applying monotonicity to $f_+$ in
\eqref{eq:app-fullset} gives the interval claim, and strict monotonicity
gives uniqueness.

For the converse, take $d=1$, $e=0$, $a=1$, $L=\kappa>1$,
$D=Q=[1,2]$. Choose distinct interior locations $s_1<\cdots<s_n$,
and sufficiently small $\delta>0$ so the intervals below are disjoint,
lie strictly inside $(1,2)$, and $s_i-\delta\geq1$.
Observe $f(s_i)=s_i-\delta$. These observations are consistent because
their pairwise slope is one. For every $q>1$ and every $i$,
\[
s_i-\delta-\kappa|q-s_i|<q.
\]
Hence $q_-=1$. The exclusion interval generated by sample $i$ is
\[
\left(s_i-\frac{\delta}{\kappa+1},\quad
      s_i+\frac{\delta}{\kappa-1}\right).
\]
There are exactly $n$ disjoint holes in $[1,2]$, leaving $n+1$
nonempty components.
\end{proof}

\paragraph{Small example.}
Strict subcriticality also makes every feasible intersection a first hit,
because each field has a unique root. At the critical value this equality
can fail through plateaus. For instance, take $Q=D=[1/4,2]$, $e=0$,
$a=L=1$, and the exact observation $f(1)=1$. The lower residual vanishes
on $[1/4,1]$, so $q_-=1$; the upper residual vanishes on $[1,2]$.
Thus the first-hit support is $[1,2]$ but the all-intersection support is
$[1/4,2]$. Connectedness at $\kappa=1$ does not assert equality of
these two supports.

With $Q=D=[1/4,2]$, $e=0$, $a=1$, $L=2$ and one exact observation
$f(1)=1/2$, the identified inverse-depth set is
\[
\mathcal Q=[1/4,5/6]\ \cup\ [3/2,2].
\]
For $z=1/q$, its depth image is $[1/2,2/3]\cup[6/5,4]$.
The interval hull includes an impossible open depth band $(2/3,6/5)$.
This example is a consequence of the model class, not a numerical failure.

\section{Shared scenes and joint depth feasibility}
\label{app:joint}

For several admissible rays querying one source field, all near endpoints
$z_j(q_{+,j})$ are attained simultaneously by the same field $f_+$;
all far endpoints $z_j(q_{-,j})$ are attained simultaneously by $f_-$.
Arbitrary choices from the separate identified sets need not be jointly
attainable. The following criterion checks the exact joint constraint.

\begin{proposition}[Exact joint first-hit membership]
\label{prop:app-joint}
Consider $m$ admissible rays $x_j(q)=e_j+a_jq$ querying the same class
$\mathcal F$, and proposed inverse depths $\widetilde q_j\in Q$.
Append exact observations
\[
 f(x_j(\widetilde q_j))=\widetilde q_j,\qquad j=1,\ldots,m.
\]
If the augmented information fails \eqref{eq:app-feasibility}, the
proposed vector is impossible. Otherwise define its minimal field
\begin{equation}
f_-^{\rm aug}(x)=\max\left\{f_-(x),
\max_j[\widetilde q_j-L\|x-x_j(\widetilde q_j)\|]\right\}.
\label{eq:app-joint-witness}
\end{equation}
The proposed vector is jointly attainable as the frontmost depths if
and only if, for every $j$,
\begin{equation}
\max\{q\in Q:f_-^{\rm aug}(x_j(q))=q\}=\widetilde q_j.
\label{eq:app-joint-check}
\end{equation}
When these checks pass, $f_-^{\rm aug}$ is a single global witness for
all proposed first hits. After original-data feasibility has been
checked, augmented feasibility and these last-root checks require
$O((mn+m^2)d)$ arithmetic operations using the lower-cone scalar solver.
\end{proposition}
\begin{proof}
By Lemma~\ref{lem:app-envelopes}, the augmented class is nonempty exactly
when its pairwise conditions hold, and then its pointwise minimum is
\eqref{eq:app-joint-witness}. That field equals $\widetilde q_j$ at
the $j$th candidate position. Thus its greatest root on ray $j$ is at
least $\widetilde q_j$. Every field in the augmented class is pointwise
at least this minimum; the endpoint-sign argument in
Theorem~\ref{thm:app-fullset} shows that its greatest root on ray $j$
is also at least the minimum field's greatest root. Hence if that
root exceeds $\widetilde q_j$, no augmented field can realize the
proposed first hit. If all equalities in \eqref{eq:app-joint-check}
hold, the minimum field itself realizes the entire vector. There are
$O(mn+m^2)$ new pairwise conditions, and each of $m$ last-root queries
uses $n+m$ lower cones. This proves the operation count.
\end{proof}

\paragraph{Why candidate-value consistency is insufficient.}
Take $Q=[1/4,2]$, $D=[-1/4,2]$, $L=3$, and the one exact source
observation $f(-1/4)=1/4$. Let two rays from the same lateral pose be
$x_1(q)=q$ and $x_2(q)=q-1/2$. The proposed vector
$(\widetilde q_1,\widetilde q_2)=(1/4,1)$ belongs to the two separate
identified sets. Its exact candidate observations
$f(1/4)=1/4$ and $f(1/2)=1$ are also Lipschitz-consistent with each
other and with the original observation. Nevertheless they are not
jointly attainable as first hits: the cone forced by $f(1/2)=1$
raises the minimum field to $1-3|x-1/2|$, whose last intersection
with ray 1 is $q=5/8>1/4$. The second candidate necessarily hides the
proposed farther first hit on the first ray.

\section{Optimal decisions from the identified set}
\label{app:decisions}

\begin{corollary}[Minimax relative depth error and exact abstention rule]
\label{cor:app-minimax}
Let $d_-=z(q_+)$ and $d_+=z(q_-)$, both strictly positive. Among all
deterministic estimators based on the same observations and assumptions,
the smallest worst-case relative target-depth error is
\begin{equation}
R^*=\frac{d_+-d_-}{d_++d_-},\qquad
\widehat d^*=\frac{2d_-d_+}{d_-+d_+}.\label{eq:app-minimax}
\end{equation}
For a proposed value $\widehat d>0$, its exact worst-case error is
\begin{equation}
R(\widehat d)=\max\left\{\frac{\widehat d}{d_-}-1,
                              1-\frac{\widehat d}{d_+}\right\}.
\label{eq:app-value-risk}
\end{equation}
For $0\leq\tau<1$, some estimate with uniformly bounded relative error
$\tau$ exists if and only if
\begin{equation}
R^*\leq\tau\quad\Longleftrightarrow\quad
\frac{d_+}{d_-}\leq\frac{1+\tau}{1-\tau}.
\label{eq:app-abstention}
\end{equation}
\end{corollary}
\begin{proof}
The two endpoint depths are attained by admissible fields with the same
information. Hence any estimator incurs at least the maximum of its
losses at these two depths. For a fixed positive estimate, relative loss
as a function of depth is maximized at one of the smallest and largest
depths in the identified set: on depths below the estimate it is
$\widehat d/d-1$, which decreases with $d$; above the estimate it is
$1-\widehat d/d$, which increases with $d$. This proves
\eqref{eq:app-value-risk} and that the two-point lower bound is exact.
An estimate outside $[d_-,d_+]$ cannot improve the optimum.
Within this interval, equalizing the decreasing and increasing endpoint
losses gives $\widehat d^*$ and $R^*$ in \eqref{eq:app-minimax}.
The equivalence in \eqref{eq:app-abstention} follows by rearrangement.
\end{proof}

Disconnected components do not improve this scalar minimax relative-loss
rule beyond the sharp interval hull; their value is retaining the exact
set of possible outcomes, for example when a downstream query asks whether
a particular depth band is possible. Abstention here is a deterministic,
model-conditional statement, not a marginal coverage or conformal-risk
statement.

\section{Sharp sampling and measurement-error limits on a scanline}
\label{app:sampling}

This section concerns a rectified lateral camera model, written
\begin{equation}
F_f(x)=x+b f(x),\qquad b>0,\qquad \kappa=bL.
\label{eq:app-lateral}
\end{equation}
For a target coordinate $v$, an intersection satisfies $F_f(x)=v$ and
has inverse source depth $q=(v-x)/b$. Thus the first hit corresponds to
the smallest source coordinate solving the equation.
Samples are at $x_i=A+ih$, $i=0,\ldots,N$, on $[A,B]$, where $B-A=Nh$.
Their intervals are $[y_i-\varepsilon,y_i+\varepsilon]$, optionally
intersected with a positive global range. Assume the admissible class
is nonempty and every considered field has an intersection in $[A,B]$.
The global-range and full-segment conditions above are one way to ensure
this. Write $\mathcal Q_v$ for the attainable first-hit inverse depths.

\begin{theorem}[Sharp noisy-grid diameter below the fold threshold]
\label{thm:app-noisy-grid}
If $\kappa<1$, then
\begin{equation}
\operatorname{diam}(\mathcal Q_v)
\leq
\max\left\{Lh+2\varepsilon,
            \frac{2\varepsilon}{1-\kappa}\right\}.
\label{eq:app-sharp-grid}
\end{equation}
This bound is sharp over consistent grid data, admissible queries, and
positive global ranges large enough not to truncate the extremizers.
For $L>0$, the two branches exchange dominance exactly at
$2b\varepsilon=(1-\kappa)h$.
When $L=0$, both branches equal $2\varepsilon$ for every $h$.
\end{theorem}
\begin{proof}
Since $\kappa<1$, each $F_f$ is strictly increasing. Take two admissible
fields $f,g$ with roots $x_f\leq x_g$, and put
$w=q_f-q_g=(x_g-x_f)/b\geq0$.
If a grid node $s$ lies between the two roots, then
\begin{align*}
f(s)-g(s)
&\geq f(x_f)-L(s-x_f)-g(x_g)-L(x_g-s)\\
&=w-L(x_g-x_f)=(1-\kappa)w.
\end{align*}
Both sample values lie in the same interval of width $2\varepsilon$,
so $w\leq2\varepsilon/(1-\kappa)$.

If no grid node lies between the roots, both lie in one grid cell
$[a,a+h]$. Using the left endpoint interval gives
\[
w\leq2\varepsilon+L(x_f+x_g-2a),
\]
and using the right endpoint gives
\[
w\leq2\varepsilon+L(2a+2h-x_f-x_g).
\]
The minimum of the two parenthesized quantities is at most $h$.
Thus $w\leq2\varepsilon+Lh$ in this case. Taking the supremum over
pairs proves \eqref{eq:app-sharp-grid}.

For sharpness in the first regime, suppose
$2b\varepsilon\leq(1-\kappa)h$, and set
\[
W=Lh+2\varepsilon,\qquad d=bW\leq h,\qquad
 y_0=C,\quad y_1=C-Ld.
\]
On one cell $[0,h]$ define the two extremal tents
\begin{align*}
f(x)&=\varepsilon+\min\{C+Lx,C-Ld+L(h-x)\},\\
g(x)&=-\varepsilon+\max\{C-Lx,C-Ld-L(h-x)\}.
\end{align*}
They are $L$-Lipschitz and satisfy the common endpoint intervals. At
$x_f=(h-d)/2$ and $x_g=(h+d)/2$ their inverse depths differ by $W$.
Moreover
\[
x_f+bf(x_f)=x_g+bg(x_g)
 =bC+b\varepsilon+\frac{(1+\kappa)(h-d)}2.
\]
Strict monotonicity makes these their first and only roots for this
common target. Choosing $C$ sufficiently large makes both fields and
an admissible global range positive. When $d=h$ the roots are the cell
endpoints; otherwise both lie in the interior. Additional grid cells
may be appended using common Lipschitz continuations.

In the second regime take
$f(x)=C-Lx+\varepsilon$ and $g(x)=C-Lx-\varepsilon$ on a sufficiently
long grid domain, with centers $y_i=C-Lx_i$. For a target whose roots
are interior,
\[
x_g-x_f=\frac{2b\varepsilon}{1-\kappa},\qquad
q_f-q_g=\frac{2\varepsilon}{1-\kappa}.
\]
A large enough positive $C$ and range accommodate these fields. More
explicitly, take a grid domain $[0,D]$ with $D\geq 2b\varepsilon/(1-\kappa)$,
choose the target so that both roots lie in the domain, and set
$Q=[(v-D)/b,v/b]$. The endpoint inequalities for both affine fields imply
that their entire ranges lie in $Q$, and $x(Q)=[0,D]$. Choosing $C$ large
makes this range positive. In the first construction the same choice with
$D=h$ works because the maximum of $f$ and the minimum of $g$ occur at
their respective roots. Thus the sharpness examples satisfy the
global-range/full-segment model as well as the sampling theorem. This
attains the second branch. The comparison of the two branches gives
the stated transition.
\end{proof}

\begin{corollary}[Exact-sample resolution and the critical boundary]
\label{cor:app-exact-grid}
For exact observations ($\varepsilon=0$), if $bL\leq1$, then
\begin{equation}
\operatorname{diam}(\mathcal Q_v)\leq Lh.\label{eq:app-exact-grid}
\end{equation}
For $0<bL<1$ the constant is attained. At $bL=1$ it is sharp as a
supremum, which need not be attained by a nondegenerate pair of fields.
Consequently the worst-case absolute inverse-depth minimax radius for
this fixed grid is $Lh/2$ (as a supremum at equality), when the range
and query allow the extremal examples.
\end{corollary}
\begin{proof}
For $bL<1$, apply Theorem~\ref{thm:app-noisy-grid} with
$\varepsilon=0$ and its first construction. When $bL=1$, every $F_f$
is nondecreasing, though a zero interval is possible. All compatible
fields have the same projected values at the grid nodes. Their first
root lies in the same cell: select the first grid node at which that
common value is at least $v$. If the first node already has value $v$,
the first root is the domain endpoint. Otherwise the preceding node's
value is strictly below $v$, and monotonicity places the first root in
that preceding cell. The two-endpoint argument in the proof above,
with zero interval width, gives \eqref{eq:app-exact-grid}.

For sharpness at equality, let $bL=1$, take $s\in[0,1)$, and use exact
endpoint values $C$ and $C-sLh$ on $[0,h]$. Let $f$ and $g$ be the
upper and lower Lipschitz tents for those two values. At target
$v=bC+(1-s)h/2$, their first roots are
\[
x_f=\frac{(1-s)h}{4},\qquad
x_g=\frac{(3+s)h}{4}.
\]
Thus their inverse-depth difference is $(1+s)Lh/2$, which approaches
$Lh$ as $s\uparrow1$. At $s=1$ the endpoint information forces the
linear field, and its projected map has a plateau; the limiting
extremal pair is not a distinct pair. Finally, for absolute error the
midpoint of the identified-set hull has worst-case error half its
diameter, while the two attained (or arbitrarily approximated) endpoints
prove the matching lower bound.
\end{proof}

The first branch in \eqref{eq:app-sharp-grid} is an interpolation limit;
the second is a measurement-error limit near a tangency. Replacing both
by $(Lh+2\varepsilon)/(1-\kappa)$ loses the exact separation. For example,
with exact samples the sharp diameter does not diverge as
$\kappa\uparrow1$. For a grid on length $B-A$, the noiseless worst-case
radius is $L(B-A)/(2N)$. With noise, a sufficient worst-case radius
$\delta$ under this bound requires both
$\delta\geq\varepsilon/(1-\kappa)$ and
$h\leq2(\delta-\varepsilon)/L$ when $L>0$. These are statements about
the specified fixed grid and function class, not lower bounds for every
adaptive sampling strategy.

\begin{theorem}[Dense exact samples need not identify a folded first hit]
\label{thm:app-fold-obstruction}
For every $b,L>0$ with $bL>1$, there exist a fixed source domain,
a fixed positive range and target, and pairs of admissible fields
$(f_h,g_h)$ that agree at every point of a uniform grid of spacing
$h\to0$, but whose first-hit inverse depths differ by a quantity
converging to $1/(2b)$. Therefore no estimator based on these grids is
uniformly consistent over the specified Lipschitz class. This is a
worst-case statement; it does not say that dense sampling fails for
every fixed scene or every folded query.
\end{theorem}
\begin{proof}
Take $D=[0,1]$, target $v=2$, and $Q=[1/b,2/b]$. Then
$x(q)=2-bq$ maps all of $Q$ onto $D$. Let
$m=\min\{bL-1,1\}>0$. For even $N\geq2$, set $h=1/N$ and
$\delta_h=mh/4$. Define a common projected baseline
\[
F_h(x)=2-\delta_h+(x-1/2)_+
\]
and a nonnegative triangular bump supported on the first cell,
\[
\phi_h(x)=
\begin{cases}
 m\min\{x,h-x\},&0\leq x\leq h,\\
 0,&h<x\leq1.
\end{cases}
\]
Put
\[
g_h(x)=\frac{F_h(x)-x}{b},\qquad
f_h(x)=\frac{F_h(x)+\phi_h(x)-x}{b}.
\]
Their slopes, wherever defined, are among
$0,-1/b,(-1+m)/b,(-1-m)/b$. Because $m\leq bL-1$, all have absolute
value at most $L$. They are continuous and hence $L$-Lipschitz.
Because $m\leq1$ and $h\leq1/2$, both fields take values in $Q$.
The bump vanishes at every grid node, so their exact sampled values
coincide.

The late field $g_h$ first reaches target $2$ at
$x_g=1/2+\delta_h$. The bumped field first reaches it at
$x_f=\delta_h/m=h/4$. Thus
\[
q_{f_h}-q_{g_h}
=\frac{x_g-x_f}{b}
=\frac{1}{2b}+\frac{(m-1)h}{4b}
\longrightarrow\frac{1}{2b}.
\]
The fields induce identical data, so any deterministic inverse-depth
estimator errs by at least half this separation on one of them.
Their depth values approach $b/2$ and $2b/3$; under relative depth
loss the corresponding two-point minimax error approaches $1/7$.
\end{proof}

\begin{proposition}[Arbitrarily small interval noise at the critical slope]
\label{prop:app-critical-noise}
At $bL=1$, arbitrarily small nonzero observation intervals can retain
macroscopic first-hit ambiguity even with arbitrarily dense sampling.
\end{proposition}
\begin{proof}
Again take $D=[0,1]$, $Q=[1/b,2/b]$, and $v=2$. Define
\[
F_0(x)=2+\min\{x-1/4,0\}+\max\{x-3/4,0\},\qquad
f_0(x)=\frac{F_0(x)-x}{b}.
\]
The field $f_0$ has slopes $0$ and $-1/b=-L$. For
$0<b\varepsilon<1/4$, the two fields $f_0+\varepsilon$ and
$f_0-\varepsilon$ remain $L$-Lipschitz and take values in $Q$.
On any grid they satisfy intervals centered at $f_0(x_i)$ with
half-width $\varepsilon$. Their projected maps are
$F_0\pm b\varepsilon$ and their first roots are respectively
$1/4-b\varepsilon$ and $3/4+b\varepsilon$.
The inverse-depth gap is $1/(2b)+2\varepsilon$, which does not vanish
as $\varepsilon\downarrow0$. The construction also belongs to every
class with a larger Lipschitz bound.
\end{proof}

\section{What the results do and do not establish}
\label{app:scope}

The exact identified set is conditional on a specified regularity class,
source sample intervals, global depth range, camera pose, and observed
surface domain. It is simultaneously valid for all completions in that
class and supplies witnesses proving that its uncertainty cannot be
reduced from the same information. It does not infer the regularity
constant, validate a monocular model's interval calibration, model pose
uncertainty, or certify geometry never visible in the source image.
The sampling impossibility results concern uniform worst cases near
visibility degeneracies. They do not contradict convergence for a fixed
scene with a positive crossing margin, nor do they establish a general
lower bound for adaptive sensing.

Classical Lipschitz extensions and set-membership estimation provide
\eqref{eq:app-lower}--\eqref{eq:app-upper}; validated first-zero algorithms
already use bounding functions and root intervals. The specific results
proved here are the full attainable first-hit set with global witnesses,
the finite epipolar scalar solver and component characterization, and
the sharp grid sampling/noise regimes. These distinctions should remain
explicit when assessing novelty against prior first-zero and rendering
methods.

\clearpage
\section{Prespecified validation protocol}
\label{app:validation-protocol}
\label{app:protocol}

This section documents the prespecified E1--E2 experiments reported in \cref{sec:experiments}.
The three-scene development diagnostic and its original 384-pixel windows
remain unchanged. The experiments use ten additional scene datasets and retain
all outcomes, including empty feasible classes, domain refusals, missing
reference values, and failures of the assumed scene class. Every result is traced to its executed record; no outcome is imputed.

\subsection{Data, split, and observation units}

Use the publicly available \texttt{perfect} versions of Middlebury 2014
\citep{scharstein2014middlebury}, with native image coordinates and the official
\texttt{calib.txt}, \texttt{im0.png}, \texttt{im1.png}, and
\texttt{disp0.pfm}/\texttt{disp1.pfm}. The dataset contains 23 physical scenes
with public reference disparity, in addition to ten scenes with hidden
ground truth. Exposure and illumination variants of one scene are not
independent scenes. We do not require the hidden-ground-truth evaluation
server. The following split was frozen before downloading the new test images
or reference maps:
\begin{center}
\begin{tabular}{p{0.17\linewidth}p{0.73\linewidth}}
\toprule
Role & Physical scenes \\
\midrule
Development, already used & Adirondack, Motorcycle, Pipes \\
Untouched test, $n=10$ & Backpack, Bicycle1, Cable, Classroom1, Couch,
Flowers, Mask, Shopvac, Sticks, Storage \\
\bottomrule
\end{tabular}
\end{center}
All named scenes use the \texttt{-perfect} suffix. No TUM RGB-D frame contributes an observation to this test. Independence is at the physical-scene level for resampling; the
protocol does not assert that ten deliberately chosen scenes form a random
sample of all indoor environments. Model pretraining overlap with public
images is not ruled out, so this is a held-out evaluation of this method,
not a claim of uncontaminated model generalization.

At native resolution, write $q=d+\mathrm{doffs}=f_x B/Z$, where the
calibration baseline $B$ and depth $Z$ use the same units. Set
$J=[\mathrm{doffs},\mathrm{doffs}+n_{\mathrm{disp}}]$ from permissible
calibration metadata. The official visualization-only bounds
\texttt{vmin}/\texttt{vmax} are not used. All ten selected scenes have
positive \texttt{doffs}; an out-of-range reference value is a reported
assumption failure, not a reason to tighten or expand $J$ after testing.

For the new test only, set the source-window width by the fixed rule
\[
 W_s=32\left\lceil
       \frac{\max\{384,n_{\mathrm{disp},s}+64\}}{32}\right\rceil.
\]
This rule depends only on public camera metadata: it accommodates the
whole epipolar segment at the full stereo baseline, including scenes
whose disparity range exceeds 384 pixels. Use every complete, disjoint
window starting at columns $0,W_s,2W_s,\ldots$ on 20 rows obtained by
rounding equally spaced positions between $0.1(H_s-1)$ and $0.9(H_s-1)$.
No window is selected by prediction error, texture, regularity, or solver
output. Record any incomplete trailing strip in the image-coverage report;
it is outside the prespecified window population, not an algorithmic
success or failure.

For baseline fraction $\beta\in\{1/4,1/2,1\}$, choose five distinct
integer target columns, evenly spaced after rounding, in
$[p,p+W_s-1-\beta n_{\mathrm{disp},s}]$, using the ceiling/floor of the
endpoints before spacing. Here $p$ is the window's left column.
The query has $e=x_t-\beta\,\mathrm{doffs}$ and $x(q)=e+\beta q$.
Every such ray has its entire $x(J)$ inside the window by construction.
If there are fewer than five distinct columns, retain each unique column
and log the shortfall; do not select a replacement window. Sample source
locations on grids of stride $h\in\{4,8,16\}$ pixels, adding the last
window endpoint when it is not on the grid.

\subsection{E1: held-out geometry and measured-stereo validation}

\paragraph{E1a: conditional digital-reference experiment.}
\begin{sloppypar}
At each sampled location with finite reference disparity, observe
$y_i=q_i+\eta_i$, where
$\eta_i\sim\operatorname{Unif}[-\varepsilon,\varepsilon]$ and
$\varepsilon\in\{0,0.5,1\}$ native disparity pixels; supply the interval
$[y_i-\varepsilon,y_i+\varepsilon]\cap J$.
The random seed is the first eight bytes of SHA-256 of a fixed UTF-8 key
string that joins a study tag with the scene, row, window, stride,
$\varepsilon$ and replicate, interpreted as an unsigned little-endian
integer; the generator is NumPy PCG64.
Use replicates $0,1,2$; they are repeated measurements, not independent
scenes. Source samples with missing reference values are absent
observations, never zero-filled depths. Use the fixed regularity bounds
$L\in\{0.25,1,4\}$ in these native coordinate units.
\end{sloppypar}

Define the digital reference to be the continuous piecewise-linear
interpolant of all finite native reference samples on a complete window.
Its first hit is computed analytically on each linear segment and with
the greatest-root convention in $q$, preserving tangent contacts.
For a complete window, membership in the assumed class is evaluated by
the global range, all supplied sample intervals, and the maximum absolute
adjacent reference slope. These membership labels are used only in
evaluation. They do not alter the supplied $L$, select windows, or permit
the algorithm to see unsampled reference values. A missing native value
makes this digital reference unavailable on the window; it is reported
separately rather than interpolated across an unobserved gap.

The primary operating cell is fixed as
$(L,h,\varepsilon,\beta)=(1,8,0.5,1)$, with all other combinations
reported as sensitivity analysis. The primary tolerance is five percent
relative target-depth error; two and ten percent are secondary tolerances.
The primary cell was chosen before new test maps were viewed. It is not
replaced by the best-performing sensitivity cell.

\paragraph{E1b: actual estimator errors and the second measured view.}
\begin{sloppypar}
Run the frozen OpenCV StereoSGBM implementation \citep{opencv2024sgbm} on the default image pair
at exactly quarter resolution, using area resampling, grayscale input,
\texttt{minDisparity=0},
$\texttt{numDisparities}=16\lceil n_{\mathrm{disp}}/64\rceil$,
\texttt{blockSize=5}, $P_1=200$, $P_2=800$,
\texttt{uniquenessRatio=10}, \texttt{speckleWindowSize=100},
\texttt{speckleRange=2}, \texttt{disp12MaxDiff=1},
\texttt{preFilterCap=31}, and \texttt{MODE\_SGBM\_3WAY}.
Use one fixed OpenCV release and record its build information. Divide fixed-point
output by 16, bilinearly resize disparity to native resolution, and
multiply by four. A native prediction is valid only when every contributing
quarter-resolution value is valid; no hole filling is performed.
The supplied bands have additive native-disparity half-widths
$\varepsilon\in\{0.5,1,2,4,8\}$, with primary value $1$.
These are empirical sensitivity bands, not guaranteed sensor intervals.
All methods receive the same resulting sample locations and bands.
\end{sloppypar}

For this estimator, repeat the digital-reference analysis without assuming
that sample bands contain the reference truth. At $\beta=1$, additionally
evaluate against the finite, positive, in-range right-view reference
value $q_t=d_1(x_t,y)+\mathrm{doffs}$ at the integer target pixel. Record
disagreement between that measured value and the source digital
first-hit reference. It can reflect discretization, source occlusion,
calibration or sensor error; it is not reclassified as a solver error or
discarded to improve coverage. No measured target view exists at
$\beta=1/4$ or $1/2$, so those cells have only the digital reference.
SGBM uses both RGB views, a fact included in its information budget;
this subexperiment is not called monocular inference.

\subsection{E2: fixed learned-depth interval interface}

Use the same untouched scenes, source windows, source sample grids,
regularity values, and target rays. Run two fixed pretrained models on the
default left RGB image, without training, scale alignment to test ground
truth, or test-time selection:
\begin{enumerate}[leftmargin=*]
\item Depth Anything V2 Metric Indoor Small \citep{yang2024dav2}, official
Hugging Face checkpoint\\
\texttt{Depth-Anything-V2-Metric-Indoor-Small-hf}.
\item ZoeDepth NYU \citep{bhat2023zoedepth}, official checkpoint
\texttt{zoedepth-nyu}.
\end{enumerate}
The checkpoint revisions were pinned from the official model repositories
before test data acquisition. Load both model and processor at the pinned
revision, in evaluation mode, float32 and batch size one. Use each saved processor's preprocessing;
return predictions to native size using its documented depth postprocessing
(including ZoeDepth's padding removal). Use one unflipped forward pass,
and declare that choice for both models. There is no training. Do not quantize predictions to an
8-bit visualization before analysis.

Convert positive finite metric predictions in meters to
$\widehat q=f_x(B/1000)/\widehat Z$.
For each fixed
$\lambda\in\{0.05,0.10,0.20,0.40,0.80\}$ supply
\[
 [\ell_i,u_i]
 =[(1-\lambda)\widehat q_i,(1+\lambda)\widehat q_i]\cap J.
\]
The primary value is $\lambda=0.20$, with $L=1$, $h=8$, and $\beta=1$.
An empty clipped interval is an inconsistent input, not an observation
whose center may be silently clipped. Nonpositive/nonfinite model output
is an absent observation and is counted. Report all fixed bands, including
near-vacuous wide bands and incompatible narrow ones. The purpose is to
test whether a useful interface exists between real model errors and the
conditional geometric theorem, not to claim that these multipliers calibrate
a desired probability of correctness.

Evaluate source pointwise band coverage, simultaneous coverage over each
window's supplied sample locations, digital-reference class membership,
and target containment. This distinguishes source prediction failure from
between-sample regularity failure and geometric propagation error. No
exchangeability is asserted for pixels or windows. In particular, a
five-scene calibration set would not support a nonvacuous standard
90\% split-conformal guarantee for a maximum scene score: its required
order statistic would be the sixth of five observations. We make no such
calibration claim and use no test-label-calibrated radius.
Windows with no supplied observations are tabulated separately; their
vacuous sample-coverage statement is not counted as evidence that a model's
bands are accurate.

\subsection{Comparisons, metrics, statistics, and claim gates}

\paragraph{Identical-information comparisons.}
Compare the exact first-hit support and its sharp hull against:
(i) the support of all feasible intersections
$\{q:\lo(e+aq)\leq q\leq\up(e+aq)\}$, which omits the first-hit
restriction;
(ii) the same first-hit calculation using only the two observations nearest
the midpoint $e+a(q_{\min}+q_{\max})/2$ (ties broken by index), which
isolates the value of nonlocal constraints;
(iii) rendering the classical source-envelope midpoint
$(\lo+\up)/2$; and
(iv) piecewise-linear and shape-preserving cubic interpolation \citep{fritsch1984pchip} through
the supplied interval centers, rendered with first-root isolation.
Point methods are evaluated for error, not described as uncertainty
certificates. Any no-root or invalid-render outcome is retained.
The exact envelope-endpoint hull is an implementation identity for our
method, not an independent competitor. No claim that disconnected support
improves the worst-case relative error of its own hull is tested: their
extrema, hence that minimax risk, coincide.

\paragraph{Accounting and outcomes.}
For each scene and fixed configuration report the number of prespecified
queries; absent observations; domain/range failures; empty classes;
missing digital and measured references; feasible finite outputs; and
verified reference membership. Report the following with their explicit
denominators:
\begin{itemize}[leftmargin=*]
\item reference containment among all finite outputs with an evaluable
reference, and separately among queries whose complete digital reference
belongs to the assumed class;
\item median and 90th percentile sharp relative uncertainty
$R^*=(d_+-d_-)/(d_++d_-)$, interval-width reduction against all-intersection
and local-information outer bounds, component counts, and the fractional
length of impossible gaps inside the target-depth interval hull;
\item acceptance fraction among \emph{all prespecified queries} at
$R^*\leq\tau$, $\tau\in\{0.02,0.05,0.10\}$, plus observed relative-error
violation frequency among accepted evaluable queries;
\item median and 90th percentile point-estimation relative error on common
evaluable queries, and solver runtime per query.
\end{itemize}
An empty acceptance set is reported as zero coverage and undefined
conditional error, never zero risk. Regularity violations remain visible
in unconditional tables; filtering to admissible references is identified
as an oracle diagnostic, not a deployable selection rule.
For floating-point containment, record the distance to the support in
inverse-depth units and use tolerance
$10^{-7}\max\{1,q_{\max}\}$; also report the largest raw residual.
Use NumPy's linear-interpolation quantile convention. Numerical tolerances
do not convert floating-point outputs into formal arithmetic certificates.

Average scene-level metrics with equal scene weights. Show every scene's
primary result. Use 10,000 paired bootstrap resamples of the ten complete
scene records (NumPy PCG64 seed 57057) for descriptive 95\% intervals.
Within a resample, retain all windows, rays and measurement replicates of
each selected scene. These intervals describe this small scene collection;
they are not formal population guarantees. Do not use millions of pixels
as independent observations. Report effect sizes and scene heterogeneity;
no significance threshold is required for a mathematically nested-set
width inequality.

Every violation of digital-reference containment when all assumptions hold
is treated as a correctness issue requiring investigation and resolution.
If memberships are rare or intervals are usually uninformative, retain the
theorems and the negative application finding, and remove practical
validity/reuse claims. If the exact support materially excludes impossible
depth bands while remaining useful on a measurable fraction of held-out
queries, the application claim is restricted to that reported model and
operating range. Successful learned-model results establish empirical
usefulness only; they do not promote a heuristic source band, an estimated
$L$, or an interpolated digital scene into a physical-scene certificate.
No temporal speedup or dynamic-scene claim is licensed by these tests.

E1a, E1b, and both E2 checkpoints were evaluated at their primary
operating cells and over the entire fixed sensitivity grid, without
replacing a primary setting. Complete counts, per-scene outcomes,
bootstrap intervals, and figures appear in \cref{app:extended-results}.
This protocol was fixed prospectively by the authors; it was not an
externally registered preregistration.

\subsection{Optional studies}
\label{app:optional}
The following are \emph{optional} extensions. No claim established in the
present paper depends on their outcomes. Both protocols were fixed before
execution; their outcomes are reported in
\cref{tab:optional} and \cref{app:reexecution}.
They do not reopen the chosen estimand or introduce a new method.

\paragraph{O1: general-pose, two-dimensional digital-scene validation.}
Use the same ten Middlebury \texttt{-perfect} left disparity maps and
calibration. Tile each native image with complete, disjoint $128\times128$
windows starting at the origin; record trailing strips and every window
containing missing reference values. For a complete window define the
source scene as the continuous bilinear interpolant of native
$r=d+\mathrm{doffs}$, with metadata range $J$ as in E1.
This is a digital scene constructed from measured data, not a measured
new target image. Its Euclidean Lipschitz constant is checked exactly
within this interpolation model: in each bilinear cell the squared gradient
norm is convex in each coordinate separately, so its maximum occurs at
a cell corner; maximize those corner norms over all cells.

Supply source grids with strides $h\in\{8,16\}$ in both dimensions,
including window edges, and reference-derived bands of half-width
$\varepsilon\in\{0,0.5,1\}$ with seeds
keyed by scene, window corner, stride, $\varepsilon$ and replicate, with
the same SHA-256-to-PCG64 conversion as E1. Use three noise
replicates and fixed $L\in\{0.25,1,4\}$ in scaled inverse-depth per native
pixel. Membership uses the full bilinear gradient bound and is an oracle
evaluation label only. The primary cell is $h=8$, $\varepsilon=0.5$, $L=1$.

Keep the source intrinsics for the artificial target camera. Let $B$ be
the physical stereo baseline in metres. For camera calculations use
physical inverse depth $q=r/(f_xB)$ and rescale the intervals and $L$
by the same factor. The primary relative pose is a
$+0.5^\circ$ yaw and translation $(0.02B,0,0)$; convention is
$X_t=RX_s+t$. Sensitivity poses are zero rotation with translations
$(\pm0.02B,0,0)$, $(0,\pm0.02B,0)$, and $(0,0,\pm0.01B)$, and zero
translation with yaw or pitch $\pm0.25^\circ,\pm0.5^\circ$. Project each
source-window center at the midpoint of the metadata inverse-depth range
into the proposed target camera and center a $5\times5$ target-ray grid
at pixel offsets $\{-8,-4,0,4,8\}^2$ around that point. This placement
uses only camera metadata and the fixed range. Every planned ray remains
in the denominator; failed orientation, positive-depth, or full-segment
checks are refusals.

For reference first hits, split the epipolar segment at every native
horizontal/vertical cell crossing. Substitution of its affine trajectory
into a bilinear cell gives a quadratic equation in inverse depth. Solve
all cell equations, retain their unsquared cell membership conditions,
include shared boundaries and root plateaus, and take the greatest root.
This independent reference does not use the solver's cone envelopes.
Compare the exact support, all-intersection support, two-nearest-sample
support, source-envelope midpoint rendering, and bilinear sample-center
interpolation on common queries. Report the E1 coverage, class-membership,
acceptance, gap, point-error, runtime and memory metrics, with descriptive
10,000-resample scene intervals. Check that every true in-class reference
hit is contained within the same $10^{-7}\max(1,r_{\max})$ tolerance and
retain the largest raw residual. For a fixed set of five central target
rays per window, the independently rendered true vector must also pass
joint membership whenever the source belongs to the class. This is a
correctness check, not an expected accuracy gain. A failure requires a
solver/reference investigation; rare admissible queries limit practical
scope and must not be removed.

\paragraph{O2: explicit pose-misspecification stress test.}
Reuse E1a's ten-scene primary source observations, all three noise replicates,
and the measured right-view queries at $\beta=1$. Keep observations,
$L=1$, $h=8$, and $\varepsilon=0.5$ fixed. Evaluate the correct calibrated
pose and 20 one-axis perturbations: translation offsets
$\{-0.05B,-0.01B,0.01B,0.05B\}$ separately along each of three axes,
and rotations $\{-0.5^\circ,-0.1^\circ,0.1^\circ,0.5^\circ\}$ separately
about yaw and pitch. Rotations are left-multiplied onto the correct
relative rotation; translation offsets are in target-camera coordinates.
Do not combine perturbations or fit their magnitudes to results.

For each assumed pose, run the full geometric admissibility checks and
convert every support through its own target-depth map $z(q)$ before
comparison to the fixed measured target depth
$f_xB/(d_1+\mathrm{doffs})$, with $B$ in metres. Report refusals, accepted fraction at
2\%, 5\%, and 10\%, containment, and observed violations among accepted
evaluable outputs. Use paired scene differences from the correct-pose
condition and the same scene-bootstrap protocol. No robust-pose guarantee
is claimed: wrong pose deliberately violates a theorem premise. The study
quantifies sensitivity, and cannot establish an error allowance by selecting
a perturbation that happens to perform well.

\emph{Executed form.} The text above reuses one-dimensional scanline
observations while applying vertical and pitch perturbations, which is not
executable under a general pose. The executed study therefore used a
two-dimensional bilinear variant with the same 21 fixed conditions. In that form every query, including the correct-pose control, was
refused by the full-segment domain check (\cref{app:reexecution}), so O2
provides no sensitivity measurement.

\begin{table*}[ht]
\centering\small
\begin{tabular}{lrrrrr}
\toprule
Study & Queries & Applicable & \shortstack{In-class\\contained} & \shortstack{Accept at 5\%\\{[}95\% interval{]}} & \shortstack{Accepted\\$>$5\% errors} \\
\midrule
\OptionalResultsRows
\bottomrule
\end{tabular}
\caption{Optional studies on the held-out data
(\cref{app:reexecution}). O1 is compute-bounded to four
complete windows per scene; its truth is a bilinear digital scene, so
containment and errors are conditional digital checks, not physical-scene
claims. Acceptance is an equal-scene mean with a descriptive 95\%
scene-bootstrap interval; containment and error columns are pooled counts.
O2 uses measured target truth under an intentionally misspecified pose; every
O2 query was refused, so it supports no sensitivity claim. O2-w is a labelled
post-hoc wide-window variant (\cref{app:reexecution}); its rows pool the
listed number of conditions over the nine evaluable scenes, give the
equal-scene acceptance mean without an interval, and count accepted queries
with measured error above 5\%. No nonprimary setting was dropped.}
\label{tab:optional}
\end{table*}

\clearpage
\section{Completed computational studies and full result accounting}
\label{app:extended-results}

\subsection{Verification of the finite solver and analytic constructions}

The numerical reference implementation and the rational oracle were written
separately. The oracle splits a one-dimensional absolute value at its rational
kink, intersects rational affine inequalities, and retains exact open/closed
endpoint semantics. It does not call the numerical solver's cone routine.
The 600 generated rational cases use seed 570026 and retain their inputs and
exact supports as rational strings. Zero slope, zero ray motion, critical
slope, tangencies, and disconnected supports are included. Numerical equality
here means the same component structure with the stated small endpoint
residual; it does not replace the real-arithmetic proof.

\begin{table}[ht]
\centering
\begin{tabular}{lrrl}
\toprule
Check & Cases/checks & Failures & Largest discrepancy or scope \\
\midrule
Independent rational support oracle & 600 & 0 & $3.464\times10^{-14}$ endpoint error \\
Generated one-dimensional fields & 500 & 0 & Reference first-hit inclusion \\
Generated two-dimensional fields & 300 & 0 & Reference first-hit inclusion \\
Constructed global witnesses & 2,064 & 0 & $1.128\times10^{-13}$ constraint residual \\
Subcritical sharpness & 72 & 0 & $8.794\times10^{-14}$ diameter error \\
Critical exact-sample sharpness & 30 & 0 & $7.319\times10^{-14}$ diameter error \\
Critical positive-noise construction & 24 & 0 & Both prescribed hits retained \\
Supercritical exact-sample construction & 12 & 0 & Both prescribed hits retained \\
Known jointly realizable fields & 300 & 0 & $2.220\times10^{-16}$ joint residual \\
Fast/reference numerical comparison & 2,400 & 0 & Includes 400 tangent fixtures \\
Incremental/reference joint comparison & 2,200 & 0 & Zero greatest-root discrepancy \\
\bottomrule
\end{tabular}
\caption{Completed numerical verification. Rows share an implementation and
some analytic constructions; they are not independent statistical replications.
The fast implementation also passes the same 600 rational cases, which are
counted only once in the table. Separate infeasible-input and domain checks
exercise refusal paths.}
\label{tab:verification}
\end{table}

The joint-membership illustration samples pairs from the two individual
supports in the fixed example of \cref{app:joint}. Of 2,000 pairs,
311 are incompatible already as sampled values, 308 additional pairs
force an earlier hit, and 1,381 pass the exact joint criterion. Thus the
visibility test supplies information beyond an ordinary Lipschitz extension
test. These fractions describe this artificial distribution only. The
check includes the rejected pair $(1/4,1)$ and its shared
lower-field first hits $(5/8,1)$.

\subsection{Development diagnostic: every planned window is retained}

The development experiment used official \texttt{Adirondack-perfect},
\texttt{Motorcycle-perfect}, and \texttt{Pipes-perfect} native disparity
maps. The final protocol was frozen before disparity values were inspected.
It specified 20 rows
equally spaced from ten to ninety percent image height, complete disjoint
384-pixel windows from column zero, five evenly spaced target rays per
window, source strides $h\in\{4,8,16\}$, inverse-depth regularity
$L\in\{0.25,1,4\}$, additive band half-widths
$\varepsilon\in\{0,0.5,1\}$, and baseline fractions
$\beta\in\{0.25,0.5,1\}$. The 81 combinations produce
170,100 query records. Target rays are continuous subpixel coordinates in
this development diagnostic. The separately frozen held-out study uses
integer target coordinates to permit measured-right comparison.

We use scaled inverse depth $q=d+\mathrm{doffs}=f_xB/Z$. A finite dense
source sample receives a band constructed to cover its reference value;
there is no learned estimator in this diagnostic. Missing source samples
are absent observations. A complete native window defines a continuous
piecewise-linear digital reference, whose exact maximum adjacent slope
determines whether that specified interpolant belongs to the class. A
window containing any missing native reference value has no such complete
digital reference and remains in the planned-query denominator. Metadata
supplies the positive global range; the range is not fitted to each patch.

\begin{table}[ht]
\centering
\small
\begin{tabular}{lrrrrrr}
\toprule
Primary development cell & Planned & Feasible & Complete ref. & In class & Misses/in class & Mean $R^*$ \\
\midrule
Adirondack & 700 & 265 & 335 & 165 & $0/165$ & $0.862\%$ \\
Motorcycle & 700 & 270 & 235 & 160 & $0/160$ & $0.943\%$ \\
Pipes & 700 & 145 & 165 & 100 & $0/100$ & $1.466\%$ \\
\midrule
Pooled & 2,100 & 680 & 735 & 425 & $0/425$ & $1.035\%$ \\
All 81 settings, pooled & 170,100 & 60,795 & 59,535 & 31,860 & $0/31,860$ & $2.091\%$ \\
\bottomrule
\end{tabular}
\caption{Development results. Primary $(L,h,\varepsilon,\beta)=(1,8,0.5,1)$.
$R^*$ is averaged only over in-class digital references. Counts of complete
references and feasible classes have different denominators and neither
implies the other. There are 109,305 refusals over all settings. The
170,100 records reuse three physical scenes and are not independent scenes.}
\label{tab:pilot}
\end{table}

At the primary cell 490 feasible outputs have a digital reference; all
contain it, including 65 references outside the prescribed class. Across
all settings 41,325 feasible outputs have references, and 125
out-of-class references are missed. Zero observed in-class misses checks
the implementation under known digital assumptions; it does not establish
those assumptions for physical surfaces. Across all settings 6,330
in-class and 13,945 feasible outputs have multiple components. Primary
outputs have one component because $L\beta=1$.

On the same 425 primary in-class references, mean relative error is
$0.3224\%$ for the minimax estimate and $0.08474\%$ for the linear
interpolant. Across all in-class sensitivity records the corresponding
means are $0.8391\%$ and $0.08462\%$. Thus minimax robustness against
all completions does not imply improved average accuracy for the selected
digital scene. The mean exact/all-intersection inverse-depth hull-width
ratio is $1.000$ at the primary setting and $0.9213$ over the full grid.
The local comparison in this pilot uses the two samples nearest the
nominal interpolant's source intersection; the held-out protocol instead
fixes the query location using the middle of $J$ before interpolation.
Those two local-baseline definitions are not pooled.

\begin{figure}[ht]
\centering
\includegraphics[width=0.98\linewidth]{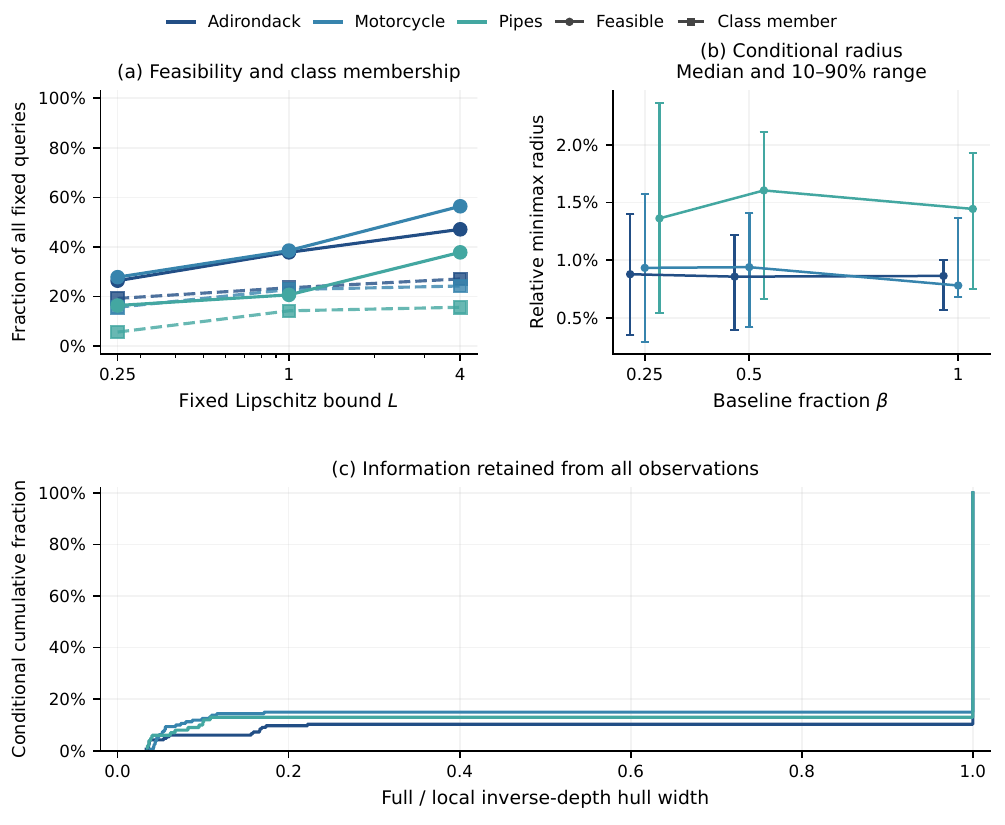}
\caption{Completed development diagnostic. Panel (a): feasible-query and
complete-reference membership rates retain every prespecified query.
Panel (b): conditional relative minimax radius, with descriptive ten-to-ninety
percent quantiles across queries, not confidence intervals. Panel (c):
full-information versus two-local-sample hull widths. Dense-reference
membership is an oracle evaluation label, not an input to a deployment
selection rule.}
\label{fig:pilot}
\end{figure}

\subsection{Held-out E1--E2 results}
\label{app:heldout-results}
The prospective design in \cref{app:protocol} was frozen before the ten
new scene images or disparity maps were accessed. The result tables below report
completed primary and sensitivity cells and preserve all missing
references, inconsistent source information, and invalid scene assumptions.
The primary operating point is never replaced after observing a sensitivity
result. Per-scene summaries are the resampling units.
All four primary experiments used the same ten prespecified physical scenes
and window/ray design. E1a includes three repeated bounded-noise realizations.
Repeated rays, samples, and measurement replicates are not independent scenes.
The official model checkpoints, source units, and native postprocessing were
independently inspected after the negative learned-band results; no unit,
relative-depth normalization, checkpoint, or padding error was found.
That check does not identify a causal explanation for the model errors.

\begin{table}[t]
\centering\small
\begin{tabular}{lrrrrrrr}
\toprule
Source & Planned & Feasible & Digital ref. & In class & Empty band & Incompatible & No obs. \\
\midrule
Reference bands & 14,700 & 5,455 & 5,250 & 4,170 & 0 & 9,245 & 0 \\
StereoSGBM & 4,900 & 2,990 & 1,750 & 265 & 0 & 1,910 & 65 \\
Depth Anything V2 & 4,900 & 1,880 & 1,750 & 225 & 2,930 & 90 & 0 \\
ZoeDepth & 4,900 & 3,155 & 1,750 & 295 & 1,560 & 185 & 0 \\
\bottomrule
\end{tabular}
\caption{Primary accounting. Empty-band and incompatible-class counts are refusals. Missing digital references remain in the planned denominator. All four primary studies have zero numerical errors, no geometric-domain refusals, and no complete-reference range failures. Each estimator has 360 queries with a complete reference violating $L$; E1a repeats these three times. Failure categories concerning the true scene can overlap, and must not be added as a partition.}
\label{tab:heldout-accounting}
\end{table}

\begin{table}[t]
\centering\small
\begin{tabular}{lrrrr}
\toprule
Source & Sample band coverage & Digital coverage & Measured coverage & In-class digital \\
\midrule
Reference bands & 878,700/878,700 & 4,530/4,530 & 5,375/5,386 & 4,170/4,170 \\
StereoSGBM & 123,880/205,640 & 1,486/1,540 & 2,678/2,932 & 265/265 \\
Depth Anything V2 & 22,705/292,900 & 252/1,155 & 299/1,839 & 225/225 \\
ZoeDepth & 79,955/292,900 & 471/1,465 & 1,128/3,089 & 295/295 \\
\bottomrule
\end{tabular}
\caption{Containment with explicit denominators. Source counts are query-weighted supplied observation incidences with finite reference values: a source sample is repeated for its five rays. They are not independent pixels. Digital and measured target coverage use finite feasible outputs with the corresponding reference, without filtering to true class membership. The final column does condition on membership. Source coverage for empty-observation windows is undefined; vacuous constraints count only when testing mathematical class membership.}
\label{tab:heldout-coverage}
\end{table}

\begin{table}[t]
\centering\small
\begin{tabular}{lrrrr}
\toprule
Source & Source coverage (\%) & In class (\%) & Accept at 5\% (\%) & Measured violations (\%) \\
\midrule
Reference bands & 100.0 [100.0, 100.0] & 23.0 [12.5, 36.0] & 30.8 [19.4, 44.3] & 0.0 [0.0, 0.0] \\
StereoSGBM & 60.8 [56.4, 64.6] & 4.1 [1.3, 7.7] & 33.8 [21.3, 46.7] & 3.7 [1.2, 6.7] \\
Depth Anything V2 & 7.3 [0.2, 17.1] & 3.9 [0.0, 9.3] & 2.8 [0.8, 5.9] & 100.0 [100.0, 100.0] \\
ZoeDepth & 25.7 [8.9, 44.7] & 4.6 [0.0, 10.1] & 9.9 [5.1, 15.8] & 89.9 [76.3, 100.0] \\
\bottomrule
\end{tabular}
\caption{Equal-scene primary rates with descriptive 95\% intervals from 10,000 paired scene resamples. The last metric excludes scenes with no accepted evaluable query: 9, 9, 8, and 10 scenes contribute, respectively. A degenerate interval at zero or one is a limitation of resampling the observed finite collection, not a statement that the population rate is known exactly. No pixel-level binomial interval is used.}
\label{tab:heldout-bootstrap}
\end{table}

\begin{table}[t]
\centering\small
\begin{tabular}{lrrrrr}
\toprule
Source & Common queries & Linear & PCHIP & Envelope midpoint & Minimax \\
\midrule
Reference bands & 4,530 & 0.0784 & 0.0844 & 0.0970 & 0.3119 \\
StereoSGBM & 1,087 & 0.4519 & 0.4550 & 0.4580 & 0.5837 \\
Depth Anything V2 & 1,155 & 25.9545 & 25.9535 & 25.9516 & 26.0897 \\
ZoeDepth & 1,465 & 23.7078 & 23.7083 & 23.6663 & 23.6683 \\
\bottomrule
\end{tabular}
\caption{Pooled mean relative-depth error in percent on exactly the same evaluable digital-reference queries for all four point estimators within a source observation condition. A valid feasible class and finite output from all comparators are required. Source conditions can have different common subsets and information budgets, so rows are not an accuracy leaderboard between the source estimators. Scene-wise median and 90th percentile errors and their bootstrap intervals will be released with the per-query records. The minimax estimator optimizes a worst-case completion loss; average point-accuracy superiority is not observed.}
\label{tab:heldout-point}
\end{table}

The fixed neural bands fail to contain most evaluable source
values and produce many empty clipped intervals. Narrow accepted target sets
can nevertheless be confidently wrong when their input assumptions fail.
For Depth Anything V2, all 169 accepted queries with measured right truth
violate the five-percent tolerance. For ZoeDepth the count is 409 of 503.
These are negative results for these checkpoints, fixed bands, and dataset,
not a theorem about all learned depth estimators. No test-label scale
alignment, band enlargement, or selection of a favorable setting replaces
the prespecified primary result.

The reference-derived bands give 4,530/4,530 digital containment on feasible
evaluable outputs, but measured right-view containment is 5,375/5,386.
Even among 4,170 digital class members, one measured right value is outside
the support. This directly illustrates why a certificate for a specified
interpolated source scene is not an unconditional certificate for physical
target measurements. At a five-percent tolerance those measured differences
do not produce an accepted point-error violation in E1a (0/5,380).

\begin{table}[t]
\centering\small
\begin{tabular}{lrrrrrr}
\toprule
Scene & Planned & Feasible & In class & Source cover (\%) & Accept at 5\% & Measured $>$5\% \\
\midrule
Backpack & 2100 & 900 & 705 & 100.00 & 891 & 0/879 \\
Bicycle1 & 2100 & 780 & 540 & 100.00 & 747 & 0/735 \\
Cable & 1500 & 560 & 450 & 100.00 & 560 & 0/560 \\
Classroom1 & 2100 & 1710 & 1530 & 100.00 & 1704 & 0/1704 \\
Couch & 900 & 150 & 75 & 100.00 & 150 & 0/150 \\
Flowers & 1200 & 340 & 255 & 100.00 & 340 & 0/340 \\
Mask & 1500 & 380 & 255 & 100.00 & 380 & 0/377 \\
Shopvac & 300 & 0 & 0 & 100.00 & 0 & -- (0) \\
Sticks & 2100 & 465 & 300 & 100.00 & 465 & 0/465 \\
Storage & 900 & 170 & 60 & 100.00 & 170 & 0/170 \\
\bottomrule
\end{tabular}
\caption{Reference bands: every primary physical scene. The final column is the number of violations divided by accepted queries with a measured right-view reference. A dash denotes an undefined error rate with zero evaluable accepted outputs, not zero error.}
\label{tab:scenes-e1a}
\end{table}

\begin{table}[t]
\centering\small
\begin{tabular}{lrrrrrr}
\toprule
Scene & Planned & Feasible & In class & Source cover (\%) & Accept at 5\% & Measured $>$5\% \\
\midrule
Backpack & 700 & 540 & 65 & 61.38 & 444 & 31/430 \\
Bicycle1 & 700 & 395 & 20 & 46.04 & 321 & 42/310 \\
Cable & 500 & 290 & 25 & 57.64 & 190 & 3/189 \\
Classroom1 & 700 & 585 & 125 & 71.72 & 480 & 0/477 \\
Couch & 300 & 120 & 0 & 60.16 & 25 & 0/25 \\
Flowers & 400 & 215 & 0 & 60.30 & 106 & 1/105 \\
Mask & 500 & 260 & 5 & 55.80 & 161 & 8/157 \\
Shopvac & 100 & 100 & 0 & 65.28 & 0 & -- (0) \\
Sticks & 700 & 315 & 20 & 63.19 & 224 & 10/223 \\
Storage & 300 & 170 & 5 & 66.78 & 70 & 0/70 \\
\bottomrule
\end{tabular}
\caption{StereoSGBM: every primary physical scene. The final column is the number of violations divided by accepted queries with a measured right-view reference. A dash denotes an undefined error rate with zero evaluable accepted outputs, not zero error.}
\label{tab:scenes-sgbm}
\end{table}

\begin{table}[t]
\centering\small
\begin{tabular}{lrrrrrr}
\toprule
Scene & Planned & Feasible & In class & Source cover (\%) & Accept at 5\% & Measured $>$5\% \\
\midrule
Backpack & 700 & 255 & 110 & 32.64 & 21 & 21/21 \\
Bicycle1 & 700 & 135 & 0 & 0.02 & 105 & 103/103 \\
Cable & 500 & 190 & 115 & 35.13 & 1 & 1/1 \\
Classroom1 & 700 & 650 & 0 & 0.37 & 8 & 8/8 \\
Couch & 300 & 5 & 0 & 0.00 & 0 & -- (0) \\
Flowers & 400 & 205 & 0 & 0.22 & 9 & 9/9 \\
Mask & 500 & 130 & 0 & 0.31 & 2 & 2/2 \\
Shopvac & 100 & 0 & 0 & 0.00 & 0 & -- (0) \\
Sticks & 700 & 75 & 0 & 0.16 & 11 & 11/11 \\
Storage & 300 & 235 & 0 & 3.86 & 14 & 14/14 \\
\bottomrule
\end{tabular}
\caption{Depth Anything V2: every primary physical scene. The final column is the number of violations divided by accepted queries with a measured right-view reference. A dash denotes an undefined error rate with zero evaluable accepted outputs, not zero error.}
\label{tab:scenes-dav2}
\end{table}

\begin{table}[t]
\centering\small
\begin{tabular}{lrrrrrr}
\toprule
Scene & Planned & Feasible & In class & Source cover (\%) & Accept at 5\% & Measured $>$5\% \\
\midrule
Backpack & 700 & 380 & 160 & 41.63 & 101 & 97/99 \\
Bicycle1 & 700 & 150 & 0 & 0.00 & 109 & 109/109 \\
Cable & 500 & 490 & 70 & 79.29 & 153 & 62/150 \\
Classroom1 & 700 & 630 & 0 & 0.30 & 2 & 2/2 \\
Couch & 300 & 35 & 0 & 0.00 & 17 & 17/17 \\
Flowers & 400 & 335 & 0 & 21.78 & 11 & 6/10 \\
Mask & 500 & 340 & 0 & 2.17 & 44 & 44/44 \\
Shopvac & 100 & 30 & 0 & 0.00 & 12 & 11/11 \\
Sticks & 700 & 510 & 65 & 64.01 & 61 & 60/60 \\
Storage & 300 & 255 & 0 & 48.32 & 1 & 1/1 \\
\bottomrule
\end{tabular}
\caption{ZoeDepth: every primary physical scene. The final column is the number of violations divided by accepted queries with a measured right-view reference. A dash denotes an undefined error rate with zero evaluable accepted outputs, not zero error.}
\label{tab:scenes-zoe}
\end{table}

\subsection{Completed full sensitivity grids}
All 3,175,200 frozen query--configuration--replicate records have been
computed, with no numerical solver failures and 553,215/553,215 digital
class-member first hits contained. These records repeatedly evaluate the
same ten physical scenes, and are not millions of independent samples.
The separate primary executions repeat cells already in the full grid;
they are not added to these totals. Every primary summary matches its
full-grid counterpart once a documented correction to one baseline
computation is applied.

\begin{table}[t]
\centering\small
\begin{tabular}{lrrrrrr}
\toprule
Source & Cells & Query records & Feasible & In-class coverage & Disconnected & Numerical failures \\
\midrule
Reference bands & 81 & 1,190,700 & 514,095 & 321,570/321,570 & 113,744 & 0 \\
StereoSGBM & 135 & 661,500 & 436,335 & 95,070/95,070 & 65,663 & 0 \\
Depth Anything V2 & 135 & 661,500 & 347,235 & 67,995/67,995 & 24,136 & 0 \\
ZoeDepth & 135 & 661,500 & 438,030 & 68,580/68,580 & 31,979 & 0 \\
\bottomrule
\end{tabular}
\caption{Entire fixed sensitivity grid, including all outcomes. E1a uses three noise realizations per query and configuration; the other interfaces use one. Counts of disconnected outputs concern feasible classes, not only oracle-selected references. Gaps below the prescribed numerical tolerance are ignored for component statistics, while raw support endpoints are retained in the per-query records.}
\label{tab:full-grid}
\end{table}

\begin{table}[t]
\centering\small
\begin{tabular}{lrrrr}
\toprule
Source & Feasible & Disconnected (\%) & Depth gaps (\%) & Hull reduction (\%) \\
\midrule
Reference bands & 8,540 & 92.1 [82.4, 97.6] & 23.6 [20.1, 25.5] & 33.7 [27.1, 39.9] \\
StereoSGBM & 3,890 & 60.0 [42.9, 74.1] & 13.9 [7.9, 19.0] & 16.9 [8.4, 25.3] \\
Depth Anything V2 & 1,955 & 52.0 [26.0, 75.7] & 1.0 [0.5, 1.6] & 9.2 [4.0, 16.0] \\
ZoeDepth & 3,305 & 47.8 [29.2, 65.5] & 0.8 [0.3, 1.3] & 7.8 [3.3, 13.9] \\
\bottomrule
\end{tabular}
\caption{Predicted folded regime, changing only $L$ from one to four relative to each primary cell ($h=8$, $\beta=1$, source band unchanged). Intervals are descriptive 95\% scene-bootstrap intervals. Disconnectedness is a mean scene fraction of feasible outputs; gap fraction and hull reduction are means of within-scene medians. Gap fraction is excluded target-depth length divided by its sharp hull length. Hull reduction is one minus the first-hit depth-hull width divided by the all-intersection depth-hull width. No query is selected by its measured error or reference class membership. All primary cells have zero gaps and zero reduction in this comparison.}
\label{tab:folded-sensitivity}
\end{table}

The folded settings expose a substantive difference between
first-hit identification and all-intersection bounds on real depth-derived
inputs. This is a statement about the supplied completion classes.
It does not repair invalid source intervals. In particular, wide learned
bands can improve observed containment while making the identified sets
uninformative. All fixed bands, including the wide $\lambda=0.8$ settings,
are retained. A cell accepting no queries has undefined observed
conditional risk, never zero risk.

\begin{figure}[p]
\centering
\includegraphics[width=\linewidth]{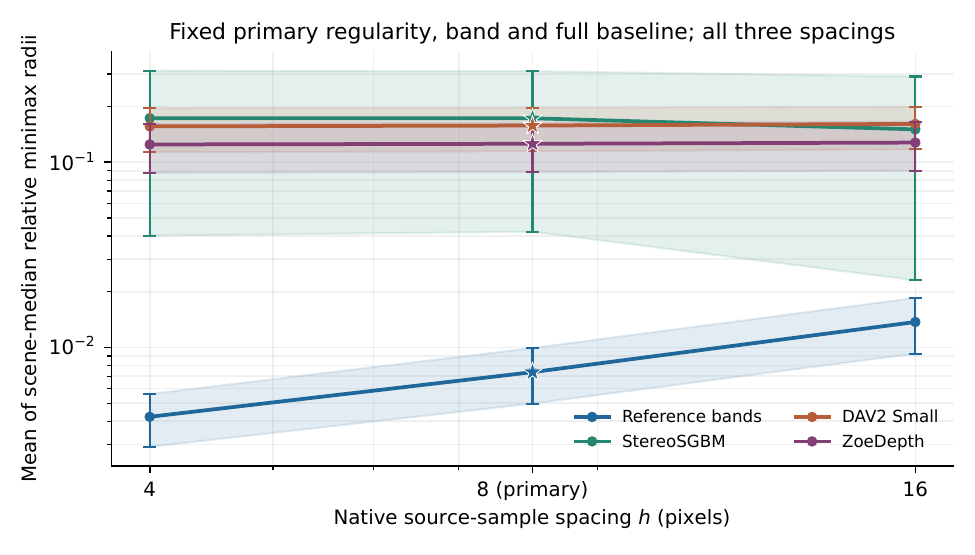}
\caption{Source spacing with each primary source band, $L=1$, and
$\beta=1$. Curves are equal-scene means of within-scene median relative
minimax uncertainty $R^*$. Shading and whiskers show descriptive 95\%
scene-bootstrap intervals; stars mark the frozen stride-eight primary
cell. The contributing scene count can depend on feasibility and is
reported in the aggregates. These are conditional uncertainty radii,
not empirical calibrated risks.}
\label{fig:heldout-spacing}
\end{figure}

\begin{figure}[p]
\centering
\includegraphics[width=\linewidth]{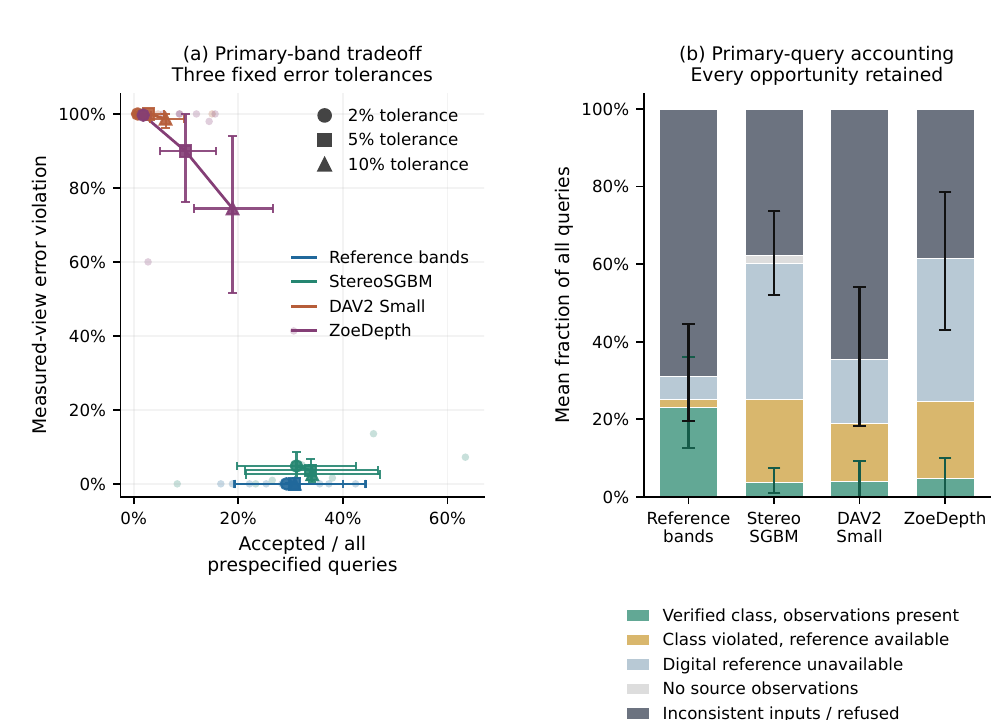}
\caption{Completed primary operating-point diagnostics. Left: equal-scene
acceptance and measured-right error-violation rates at tolerances
2\%, 5\%, and 10\%, with descriptive scene-bootstrap intervals. Faint
points show individual scenes at 5\%; empty-acceptance risks are absent.
Right: a mutually exclusive partition of every planned query, averaged
over ten scenes. Empty observation sets have their own category; they
are not evidence of valid model bands. Green whiskers quantify nonempty
verified membership and black whiskers feasible-output fraction.}
\label{fig:heldout-accounting}
\end{figure}

\begin{figure}[p]
\centering
\includegraphics[width=\linewidth]{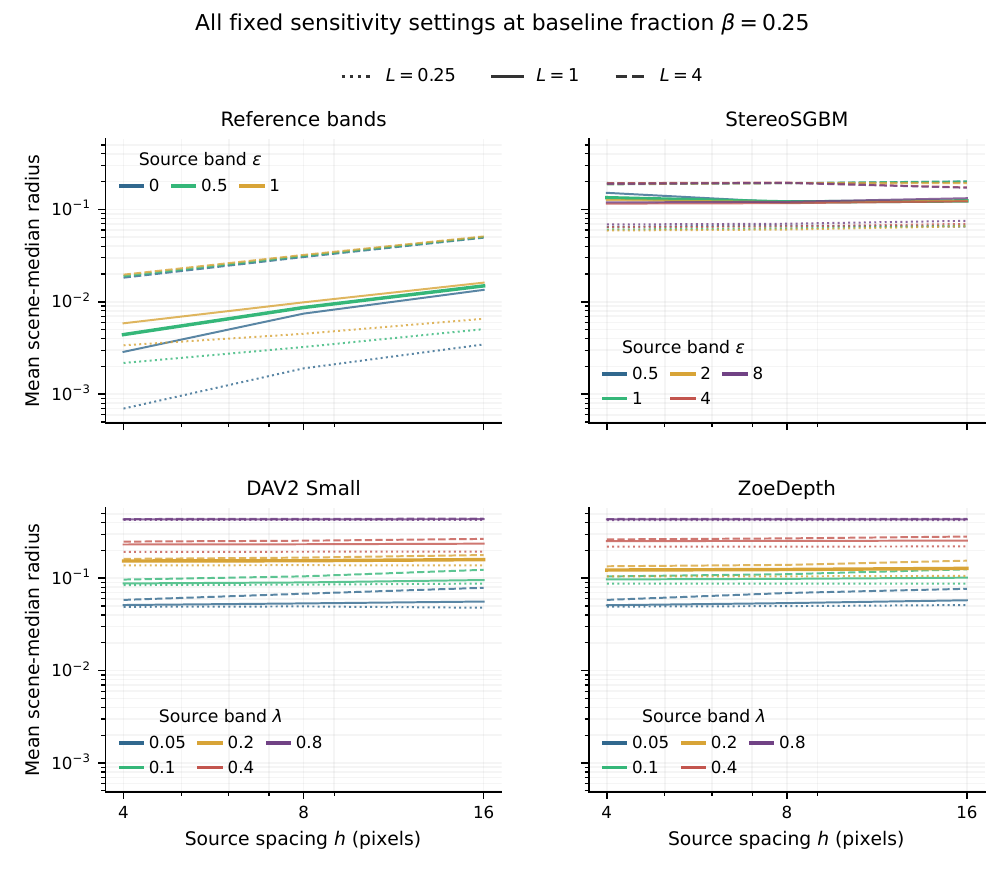}
\caption{Entire frozen sensitivity grid at $\beta=1/4$. The four panels
are the four source interfaces. Colors encode every fixed source band
and line style the regularity bound. Each line shows the equal-scene mean
of within-scene median $R^*$ among finite outputs as source stride changes.
Undefined conditional metrics remain gaps. No setting is selected by target
error. All scene counts and bootstrap intervals are in the numerical records.}
\label{fig:heldout-sensitivity}
\end{figure}

\begin{figure}[p]
\centering
\includegraphics[width=\linewidth]{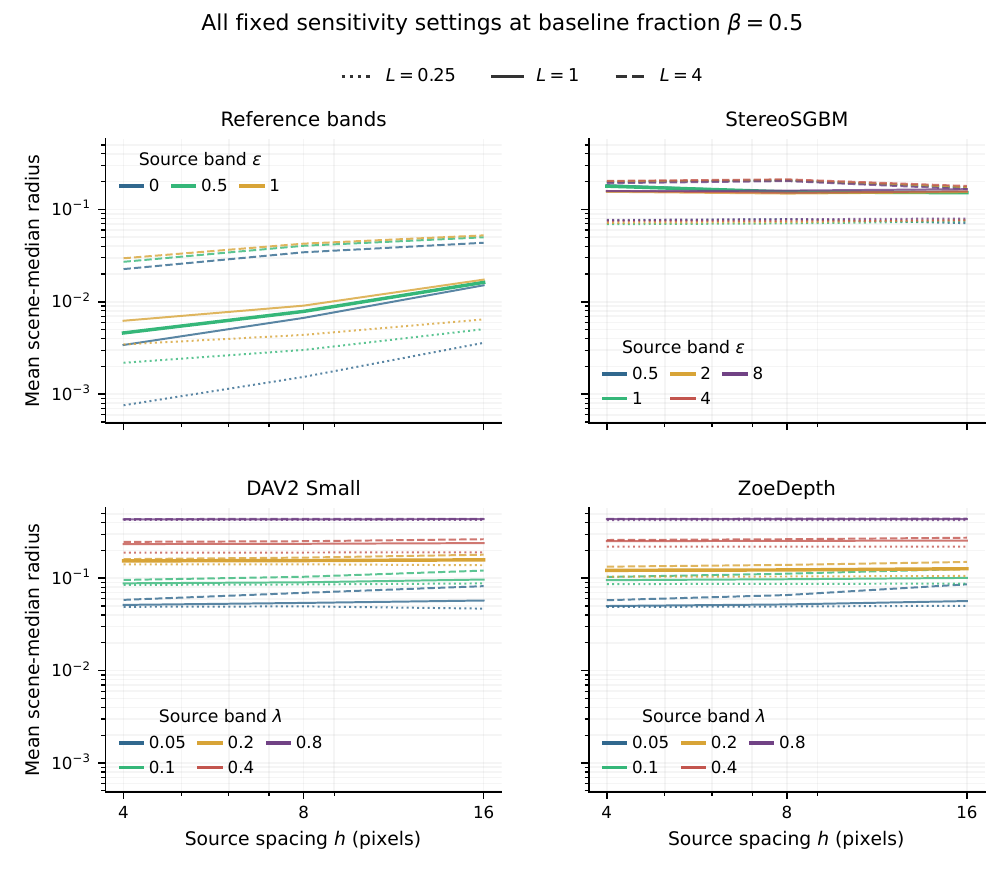}
\caption{Entire frozen sensitivity grid at $\beta=1/2$, with the same
panels, bands, regularity values, source spacings, and aggregation rule as
\cref{fig:heldout-sensitivity}. These are completion-class uncertainty
radii, not calibrated physical risks. All fixed settings are retained.}
\label{fig:heldout-sensitivity-half}
\end{figure}

\begin{figure}[p]
\centering
\includegraphics[width=\linewidth]{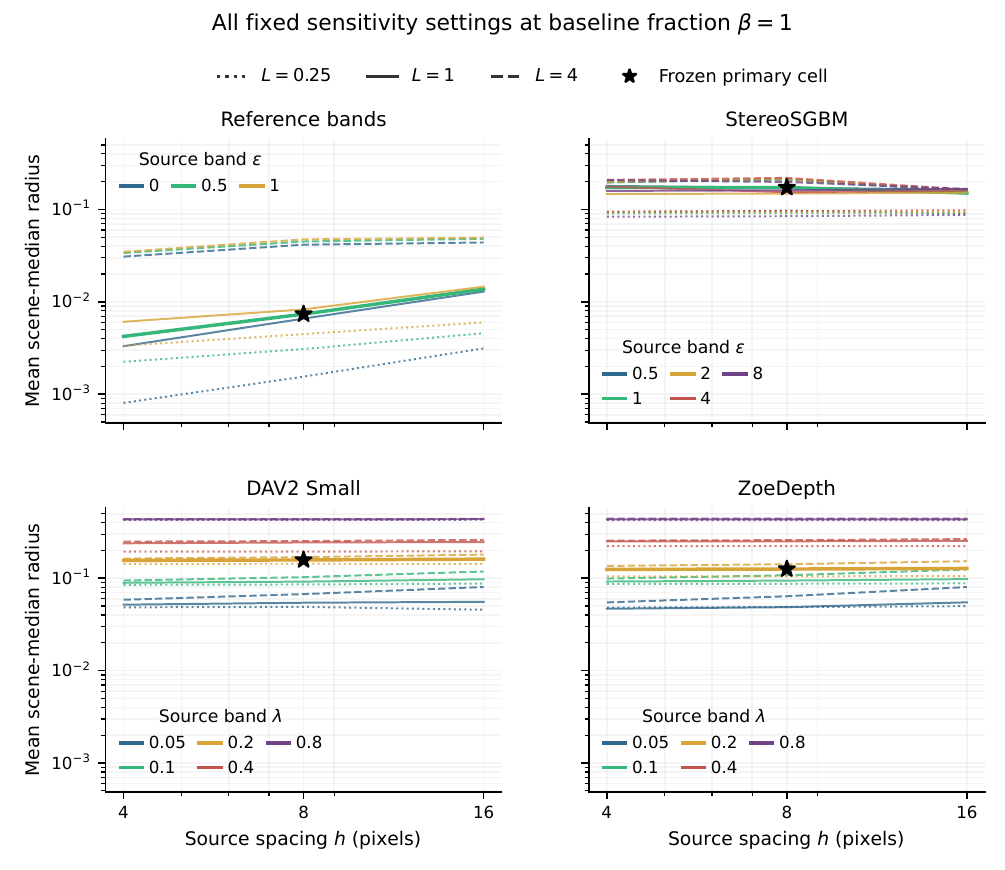}
\caption{Entire frozen sensitivity grid at $\beta=1$. Black stars mark
the frozen primary cells. Colors encode source bands and line styles the
regularity bounds, as in \cref{fig:heldout-sensitivity}. Primary-spacing
intervals appear in \cref{fig:heldout-spacing}; every full-grid cell's
interval and contributing scene count is in the numerical records.}
\label{fig:heldout-sensitivity-full}
\end{figure}

\clearpage
\section{Independent re-execution and optional studies}
\label{app:reexecution}
This section reports an independent re-execution of the frozen learned-depth
study (E2) and the executed optional studies. Every neural model ran at fp32
with batch size one, as E2 specifies, with models loaded at their pinned
revisions; the Middlebury inputs matched the recorded input hashes. Every
deviation from the prespecified settings was recorded before the affected
stage executed; except for the labelled post-hoc O2-w variant below, none
was chosen after seeing that stage's outcome. Stages interrupted for
technical reasons were rerun unchanged, and no interrupted run contributes
a reported number.

\paragraph{Archive replay.} An integrity replay re-verified \ArchiveRecords{}
frozen query/configuration records and \ArchiveMembers{} contained digital
class-member first hits. It found no numerical errors and no failures in
\ArchiveChecks{} record-integrity checks.

\paragraph{Frozen E2 re-execution.} For both learned interfaces, the
fresh records were matched one to one with the frozen records on scene, row,
window, target pixel, $L$, spacing, band, $\beta$ and replicate.
\begin{table}[!ht]
\centering\small
\begin{tabular}{lrrrr}
\toprule
Model & Records & Decision mismatches & Max.\ rel.\ difference & Acceptance flips \\
\midrule
Depth Anything V2 & \EtwoRecords & 0 & \EtwoMaxRelDAv & 0 \\
ZoeDepth & \EtwoRecords & 0 & \EtwoMaxRelZoe & 0 \\
\bottomrule
\end{tabular}
\caption{Frozen E2 versus the fresh fp32 re-execution. Decision fields include
membership, feasibility, refusal reason and component count. The relative
difference covers support endpoints, radius and the reference values.
Acceptance flips count changes at the 2, 5 and 10\% tolerances. Every entry
of \cref{tab:heldout-primary} for these two models is therefore reproduced
exactly.}
\label{tab:e2-reexecution}
\end{table}
The re-execution also recomputes StereoSGBM disparities and evaluates them
at band $0.2$. Frozen E2 used fixed StereoSGBM predictions at band $1$, so
that row is a different setting, not a re-execution, and is not interpreted
here.

\paragraph{Depth Pro (extension).} The re-execution adds Depth Pro
\citep{bochkovskii2025depthpro}, which is not part of the frozen E2, and runs
the E2 primary protocol for it. All its computation is fp32; an alternative
placement of its weights agreed with the one used to within
$5.9\times10^{-7}$ relative, which is fp32 rounding. \DepthProText{}

\paragraph{O1: general pose.} The executed configuration evaluates four
complete windows per scene, out of roughly 130 available, at every
prespecified factor level: 15 poses, two spacings, three band half-widths,
three slope bounds and three noise replicates, each with 25 target rays.
That gives \OoneQueries{} planned queries. All \OoneMembers{} evaluable
digital class members were contained. The \OoneJointChecks{} joint
first-hit checks had \OoneJointFailures{} class-member failures. No accepted
output had an error above 5\% against the digital truth
(\cref{tab:optional}). These are conditional checks of the solver against a
bilinear digital scene built from measured data. They are not
measured-target or physical-scene evidence, and the four-window cap makes
the study smaller than the uncapped protocol.

\paragraph{O2: pose misspecification.} The executable two-dimensional form
(\cref{app:optional}) ran all \OtwoConditions{} fixed conditions on \OtwoRecords{}
records. Every record, including the correct-pose control, was refused with
\emph{full epipolar segment is not observed}. The refusal does not depend on
pose error. Under this protocol, the observed source windows do not cover the
epipolar segment that each target ray's depth range sweeps. O2 therefore
measures no pose sensitivity.

\paragraph{O2-w: wider windows (post hoc, exploratory).} After seeing the O2
refusals we specified one labelled post-hoc variant before running it. It
changes only the window: its width follows E1's metadata rule,
$32\lceil\max(384,\,n_{\mathrm{disp}}+64)/32\rceil$ pixels, so that a window
can contain the full epipolar segment; the height stays 128\,px and
candidate windows are tiled every 32\,px, with at most four per scene chosen
as before. Poses, perturbations, noise, $L$, spacing, bands and metrics are
unchanged. Shopvac has no fully observed window at this height and
contributes no queries. \OtwoWText{} In every scene and condition the
accepted fraction equals the applicable fraction: the sources are ground-truth
disparities with $\pm0.5$\,px reference bands, and no feasible output had a
radius above 5\%. Wrong-pose conditions violate a theorem
premise, so these are sensitivity measurements, not certified error
allowances, and O2-w is never a substitute for O2.

\paragraph{TUM RGB-D freiburg3 (extension).} The scene-disjoint study on
TUM RGB-D \citep{sturm2012tum} is not part of the prespecified claim set.
Depth holes in the TUM reference are recorded as missing error values.
\FreiburgText{}

All numbers in this section and in \cref{tab:optional} are generated
directly from the run records.

\clearpage
\section{Motivating construction: aliasing between exact samples}
\label{app:legacy}

This section gives the construction cited in \cref{sec:intro}. It
shows why a discrete interval-propagation scheme, which bounds depth only at
sampled points and then splats those bounds into the target view, cannot
certify the first visible surface between samples. It is a constructed
example, not a prevalence estimate.

Such a scheme propagates source depth and pose intervals into projected
point boxes, then uses raster coverage and a windowed z-buffer to accept
target pixels. Its componentwise point-projection bound is valid in real
arithmetic, but a continuous-surface enclosure does not follow from bounds
at sampled points and discrete splat coverage: continuity alone does not
control between-sample geometry.

An explicit physical construction uses
\[
d(u,v)=1+\sin^2(\pi u),\quad
X_s=d(u,v)((u-c_x)/f_x,(v-c_y)/f_y,1),
\]
with $f_x=f_y=100$, a $32\times24$ source grid, principal point
$(15.5,11.5)$, and exact lateral translation $0.005$ m. All integer source
pixels have exact depth one. The target projection is
$F(u,v)=(u+0.5/d(u,v),v)$; its horizontal derivative is at least
$0.1256402784>0$. Thus the projection is smooth and one-to-one, with no
occlusion, disocclusion, or depth discontinuity at the tested target.
At target row 12, column 16, the true depth is $1.6587695728$.
The interval splat accepts $[1,1]$ when its source error radius is zero
and emits depth one, a $39.714\%$ relative error. It also accepts narrow
incorrect intervals at source radii $0.02$ and $0.05$.

The class of \cref{sec:model} makes the missing assumption
explicit. For the same surface, the inverse-depth range is $[1/2,1]$, and
its exact global slope bound is
\[
L=\frac{2\pi\sqrt{t(1-t)}}{(1+t)^2}=1.7487194432,
\qquad t=(5-\sqrt{17})/4.
\]
The first-hit solver gives inverse-depth support $[0.5335155192,1]$, hence
target depth hull $[1,1.8743597216]$, containing the truth. Its optimal
relative error bound is $0.3041928660$, so it correctly refuses a $25\%$
guarantee. This demonstrates the change in the meaning of a certificate,
not an empirical accuracy gain.

\clearpage
\section{Reproducibility, implementation, and asset provenance}
\label{app:reproducibility}

\subsection{Coordinate units and scene interpretation}

The main geometric formulation uses inverse optical-axis depth. The
Middlebury experiments use its constant scaled version
$r=\gamma q$, where $\gamma=f_xB$ in consistent camera units and
$r=d+\mathrm{doffs}$. This does not change the analysis: if
$x(q)=e+aq$, then $x(r)=e+(a/\gamma)r$, the regularity bound becomes
$L_r=\gamma L$, and $L_r\|a/\gamma\|=L\|a\|$. The target-depth map is
$z(r)=(\gamma/r-c_z)/v_z$. For rectified stereo $c_z=0,v_z=1$, a common
positive scale cancels from relative-depth error and the minimax radius.
Thus the numerical $L$ values are disparity pixels per native image pixel,
not metric inverse-depth slopes. Camera intrinsics are applied to the
image coordinates before using the regularity bound.
The sampling proof uses $v=x+bf(x)$ with $b>0$, hence $x(q)=v-bq$.
The experimental left-to-right convention has $x(q)=e+\beta q$.
Reflecting the source and target axes maps the latter to the former;
the relevant parameter is $\kappa=L|b|=L\|a\|$. Source spacing,
interval width, and the diameter bound are invariant under this reflection.

Every conditional digital-reference statement concerns one declared
scanline window and its continuous piecewise-linear surface. Such a window
can pass every digital assumption while a physical target pixel disagrees,
because the interpolant, camera calibration, or exclusion of outside-patch
occluders is imperfect. The measured-right analysis reports this distinction.
It does not use target truth to certify patch selection. Arbitrary unseen
foreground surfaces cannot be excluded by this algorithm.

\subsection{Implementation and code release}

The code will be released upon publication. It comprises the reference
first-hit solver (envelopes, witnesses and the joint-membership test), a
vectorized one-dimensional variant for large fixed grids, an incremental
joint test achieving $O((mn+m^2)d)$ after source feasibility, the
independent exact rational one-dimensional support oracle, a known-pose
pinhole adapter with the positivity, domain and target-depth decisions,
the solver and construction checks of \cref{tab:verification}, the
development diagnostic, the held-out evaluation with all refusals, scene
summaries and bootstrap intervals, and the programs that download the
frozen test scenes (recording file hashes) and run the fixed estimators.
The estimator programs never read reference disparity. Per-query records,
fixed configurations and input hashes will be released with the code.

The floating solver uses double precision, a small documented outward
padding proportional to sixteen machine epsilons, and unsquared cone
predicates. Its reported sets are numerical approximations; directed-rounding
formal certificates are not implemented. The independent rational oracle
has exact endpoints for rational one-dimensional inputs.
The incremental joint routine avoids rechecking original--original sample
pairs; it matches the reference on 2,000 candidate-pair and 200 known
two-dimensional shared-field cases.
The calibrated-camera adapter implements the source-to-target pose convention,
the distinct intrinsic transformations of $e$ and $a$, inverse-depth unit
rescaling, and target-depth minimax decisions. Checks against explicit
three-dimensional projection cover five rotations and two inverse-depth
scales, ten constant-scene solves, and the orientation/positive-depth
refusal cases. They are implementation checks, not general-pose real-scene
experiments.

Middlebury PFM decoding follows the provider's convention
\citep{middleburyPfm}: the sign of the scale header determines byte order,
and its absolute magnitude is ignored.

\subsection{Computational cost}

No neural model is trained. Mean reference-solver time was 0.962 ms per
feasible query over the development grid and 0.679 ms at the primary cell.
Vectorizing the scalar cone tests gave a 4.02-fold throughput improvement
in a fixed solver comparison, with no change to the support; this
microbenchmark is not an end-to-end view-synthesis speedup, and nothing in
this paper establishes a practical temporal-reuse speedup.

\subsection{Sources and licenses}

The paper cites primary sources for the public Middlebury dataset,
Depth Anything V2, ZoeDepth, Depth Pro, TUM RGB-D and OpenCV. Benchmark RGB
and disparity maps and pretrained weights are downloaded from their official
sources and are not redistributed. Public availability alone is not treated
as permission to relicense them. ZoeDepth's pinned model card declares MIT;
the pinned Depth Anything metric model card does not establish a license
field. Any upstream term not established from the recorded metadata remains
an upstream requirement.

\end{document}